\documentclass[11pt]{article} 
\usepackage{url}
\usepackage{smile}
\usepackage{graphicx} 
\usepackage[ruled,vlined]{algorithm2e}
\usepackage{todonotes}
\usepackage{epstopdf}
\usepackage{wrapfig}
\usepackage[colorlinks, linkcolor=blue, anchorcolor=blue, citecolor=blue]{hyperref}
\usepackage[margin=1in]{geometry}
\usepackage[normalem]{ulem}
\usepackage[export]{adjustbox}
\usepackage{mathtools, cuted}
\usepackage{natbib}
\usepackage{enumerate}
\usepackage{microtype}
\usepackage{graphicx}
\usepackage{booktabs} 
\usepackage{smile}
\usepackage{wrapfig}
\usepackage{xspace}

\usepackage{kpfonts}
\DeclareMathAlphabet{\mathsf}{OT1}{cmss}{m}{n}
\SetMathAlphabet{\mathsf}{bold}{OT1}{cmss}{bx}{n}

\newcommand{\pcrf}{{\sf AutoNER}\xspace}
\newcommand{\lrcrf}{{\sf LRNT}\xspace}
\newcommand{\kalm}{{\sf KALM}\xspace}
\newcommand{\connet}{{\sf ConNET}\xspace}

\newcommand{\ours}{{\sf BOND}\xspace}

\newcommand{\bands}{{\sf BOND}\xspace}

\begin{document}

\title{\ours: BERT-Assisted Open-Domain Named Entity Recognition with Distant Supervision}

\author{Chen Liang$\dagger$, Yue Yu$\dagger$, Haoming Jiang$\dagger$, Siawpeng Er, Ruijia Wang, Tuo Zhao, Chao Zhang \thanks{All authors are affiliated with Georgia Institute of Technology. $\dagger$ indicates equal contribution. Emails: \{\tt cliang73,yueyu,jianghm,ser8,rwang,tourzhao,chaozhang\}@gatech.edu.}}
\date{}

\maketitle


\begin{abstract}
We study the open-domain named entity recognition (NER) problem under distant supervision. The distant supervision, though does not require large amounts of manual annotations, yields highly incomplete and noisy distant labels via external knowledge bases. To address this challenge, we propose a new computational framework -- \ours, which leverages the power of pre-trained language models (e.g., BERT and RoBERTa) to improve the prediction performance of NER models. Specifically, we propose a two-stage training algorithm: In the first stage, we adapt the pre-trained language model to the NER tasks using the distant labels, which can significantly improve the recall and precision; In the second stage, we drop the distant labels, and propose a self-training approach to further improve the model performance. Thorough experiments on 5 benchmark datasets demonstrate the superiority of \ours over existing distantly supervised NER methods. The code and distantly labeled data have been released in \url{https://github.com/cliang1453/BOND}.

\end{abstract}
\section{Introduction}


Named Entity Recognition (NER) is the task of detecting mentions of real-world
entities from text and classifying them into predefined types (e.g.,
locations, persons, organizations). It is a core task in knowledge extraction
and is important to various downstream applications such as user interest
modeling \citep{karatay2015user}, question answering \citep{khalid2008impact}
and dialogue systems \citep{bowden2018slugnerds}. Traditional approaches to
NER mainly train statistical sequential models, such as Hidden Markov Model
(HMM) \citep{zhou2002named} and Conditional Random Field (CRF)
\citep{lafferty2001conditional} based on hand-crafted features.  To alleviate the burden of designing hand-crafted features, deep learning models
\citep{ma2016end,huang2015bidirectional} have been proposed for NER
and shown strong performance.  However, most deep learning methods rely on
large amounts of labeled training data. As NER tasks require token-level
labels, annotating a large number of documents can be expensive,
time-consuming, and prone to human errors. In many real-life scenarios, the
lack of labeled data has become the biggest bottleneck that prevents deep
learning models from being adopted for NER tasks.

To tackle the label scarcity issue, one approach is to use distant
supervision to generate labels automatically. In distant supervision, the
labeling procedure is to match the tokens in the target corpus with
concepts in knowledge bases (e.g.
Wikipedia\footnote{\url{https://www.wikipedia.org/}} and
YAGO\footnote{\url{https://www.mpi-inf.mpg.de/departments/databases-and-information-systems/research/yago-naga/yago/}}),
which are usually easy and cheap to access.  Nevertheless, the labels generated
by the matching procedure suffer from two major challenges. The first challenge
is \emph{incomplete annotation}, which is caused by the limited coverage of
existing knowledge bases. Take two common open-domain NER datasets as examples. From Table~\ref{tab:comp_distantlabel}, we find that the coverage of tokens on both datasets is very low (less than 60\%).This issue renders many entities mentions unmatched and produces many false-positive labels, which can hurt subsequent NER model training significantly.  The second challenge is \emph{noisy annotation}. The annotation is often noisy due to the labeling ambiguity -- the
same entity mention can be mapped to multiple entity types in the knowledge
bases.  For instance, the entity mention '\emph{Liverpool}' can be mapped to
both '\emph{Liverpool City}' (type: \texttt{LOC}) and '\emph{Liverpool Football
Club}' (type: \texttt{ORG}) in the knowledge base.  While existing methods
adopt label induction methods based on type popularity, they will potentially
lead to a matching bias toward popular types.
Consequently, it can lead to many false-positive samples and hurt the
performance of NER models. What's worse, there is often a trade-off between the label accuracy and coverage: generating the high-quality label requires setting strict matching rules which may not generalize well for all the tokens and thus reduce the coverage and introduce false-negative labels. On the other hand, increasing the coverage of annotation suffers from the increasing number of incorrect labels due to label ambiguity. From the above, it is still very challenging to generate high-quality labels with high coverage to the target corpus.

Several studies have attempted to address the above challenges in
distantly-supervised NER.  To address the label incompleteness issue, some
works adopt the partial annotation CRFs to consider all possible labels for
unlabeled tokens~\citep{yang2018distantly,shang2018learning}, but they still
require a considerable amount of annotated tokens or external tools.  To
address the label noise issue, \citeauthor{ni2017weakly} \cite{ni2017weakly}
use heuristic rules to filter out sentences with low matching quality. However,
this filtering strategy improves the precision at the expense of lowering the
recall.  \citeauthor{cao2019low} \cite{cao2019low} attempt to induce labels for
entity mentions based on their occurrence popularity in the concept taxonomy,
which can suffer from labeling bias and produce mislabeled data.  Moreover,
most of the methods mainly focus on NER tasks in specific domains (e.g.  biomedical, chemistry, etc.) where the ambiguity of the named entity is very low. When the matching ambiguity issue is more severe, such methods will be less effective especially under open-domain scenarios. Till now, training \emph{open-domain} NER models with distant
supervision remains a challenging problem.

We propose our model \bands, short for \textbf{B}ERT-Assisted \textbf{O}pen-Domain \textbf{N}amed entity recognition with \textbf{D}istant Supervision, which learns accurate named entity taggers from distant
supervision without any restriction on the domain or the content of the
corpora. To address the challenges in learning from distant supervision, our approach leverages the power of pre-trained language
models (e.g., ELMo \citep{peters2018deep}, BERT
\citep{devlin2018bert}, XLnet \citep{yang2019xlnet}) which are particularly
attractive to this task due to the following merits: \emph{First}, they are very large neural networks trained  with huge amounts of unlabeled data in
a \emph{completely unsupervised manner}, which can be cheaply obtained; \emph{Second}, due to their massive sizes (usually
having hundreds of millions or billions of parameters), they have \emph{strong
expressive power} to capture general semantics and syntactic information
effectively. These language models have achieved state-of-the-art performance
in many popular NLP benchmarks with appropriate fine-tuning
~\citep{devlin2018bert,liu2019roberta,yang2019xlnet,Lan2020ALBERT,raffel2019exploring}, which demonstrates their strong ability in modeling the text data.

To fully harness the power of pre-trained language models for tackling the two
challenges, we propose a two-stage training framework. In the first stage, we
fine-tune the RoBERTa model~\citep{liu2019roberta} with distantly-matched labels
to essentially transfer the semantic knowledge in RoBERTa, which will improve the quality of prediction induced from distant
supervision. 
It is worth noting that we adopt early stopping to prevent the model from overfitting to the incomplete annotated labels\footnote{Here the incomplete annotated labels refer to tokens wrongly labeled as type '\texttt{O}'.} and significantly improve the recall. Then we use the RoBERTa model to predict a set of pseudo soft-labels for all data. In the second stage, we replace the distantly-matched labels with  the  pseudo soft-labels and design a \emph{teacher-student} framework to further improve the recall. The \emph{student} model is first  initialized by the  model learned in the first stage and trained using  pseudo soft-labels.
Then, we update the \emph{teacher} model from the \emph{student} model in the previous iteration to generate a new set of pseudo-labels for the next iteration to continue the training of the \emph{student} model.  
This \textit{teacher-student} framework enjoys the merit that it progressively improves the model confidence over data. In addition, we select samples based on the prediction confidence of the \emph{student} model to further improve the quality of  soft labels. In this way, we can better exploit both the knowledge base information and the language models and improve the model fitting.

Our proposed method is closely related to low-resource NER and semi-supervised learning. We discuss more details in Section 5. We summarize the key contributions of our work as follows:

\vspace{0.05in}
\noindent $\bullet$ We demonstrate that the pre-trained language model can also provide additional semantic information during the training process and reduce the label noise for distantly-supervised named entity recognition. To the best of our knowledge, this is the first work that leverages the power of pre-trained language model for open-domain NER tasks with distant supervision.

\vspace{0.05in}
\noindent $\bullet$ We design a two-stage framework to fully exploit the power of language models in our task. Specifically,  we refine the distant label iteratively with the language model in the first stage and improve the model fitting under the teacher-student framework in the second stage, which is able to address the challenge of noisy and incomplete annotation.

\vspace{0.05in}
\noindent $\bullet$ We conduct comprehensive experiments on 5 datasets for named entity recognition tasks with distant supervision. Our proposed method significantly outperforms state-of-the-art distantly supervised NER competitors in all 5 datasets (4 of which by significant margins). 

\section{Preliminaries}
\newcommand{\eg}{\emph{e.g.}\xspace} 

We briefly introduce the distantly-supervised NER problem and the pre-trained language models.

\subsection{Distantly Supervised NER}

NER is the process of locating and classifying named entities in text into
predefined entity categories, such as person names, organizations, locations,
etc. Formally, given a sentence with $N$ tokens $\bX=[x_{1}, ...,
x_{N}]$, an entity is a span of tokens $\bs = [x_i, ...,x_j] \  (0 \leq
i \leq j\leq N)$ associated with an entity type.  Based on the \texttt{BIO}
schema~\citep{li2012joint}, NER is typically formulated as a sequence labeling
task of assigning a sequence of labels $\bY = [y_{1}, ..., y_{N}]$ to the
sentence $\bX$. Specifically, the first token of an entity mention with type
\texttt{X} is labeled as \texttt{B-X}; the other tokens inside that entity
mention are labeled as \texttt{I-X}; and the non-entity tokens are labeled as
\texttt{O}. 

For (fully) supervised NER, we are given $M$ sentences that are already
annotated at token level, denoted as $\{(\bX_m,\bY_m)\}_{m=1}^M$. Let $f(\bX;\theta)$ denote an NER model, which can compute $N$ probability simplexes
for predicting the entity labels of any new sentence $\bX$, where $\theta$ is
the parameter of the NER model. We train such a model by minimizing the following loss over $\{(\bX_m,\bY_m)\}_{m=1}^M$:
\begin{align}\label{supervised-NER}
\hat\theta = \argmin_{\theta} \frac{1}{M}\sum_{m=1}^\text{M} \ell(\bY_m, f(\bX_m; \theta)),
\end{align}
where $\ell(\cdot, \cdot)$ is the cross-entropy loss. 

For distantly-supervised NER, we do not have
access to well-annotated true labels, but only \emph{distant labels} generated
by matching unlabeled sentences with external gazetteers or knowledge bases (KBs).
The matching can be achieved by string matching
\citep{giannakopoulos-etal-2017-unsupervised}, regular expressions
\citep{DBLP:journals/corr/Fries0RR17} or heuristic rules 
(e.g., POS tag constraints). Accordingly, we learn an NER model by minimizing Eq. \eqref{supervised-NER} with
 $\{\bY_m\}_{m=1}^M$ replaced by their distantly labeled counterparts. 

\vspace{1ex} \noindent \textbf{Challenges.}
The labels generated by distant supervision are often noisy and incomplete.
This is particularly true for open-domain NER where there is no restriction on
the domain or the content of the corpora. 
\citeauthor{DBLP:journals/corr/Fries0RR17} \cite{DBLP:journals/corr/Fries0RR17} and \citeauthor{giannakopoulos-etal-2017-unsupervised} \cite{giannakopoulos-etal-2017-unsupervised}
have proposed distantly-supervised NER methods for specific domains
(\eg, biomedical domain), where the adopted domain-specific
gazetteers or KBs are often of high matching quality and yield high precision and high recall distant labels.
For the open domain, however, the quality of the distant labels is much worse,
as there is more ambiguity and limited coverage over entity types in open-domain KBs. Table \ref{tab:comp_distantlabel} illustrates the matching quality
of distant labels on the open-domain and the biomedical-domain datasets. As can be seen, the distant labels for the open-domain datasets suffer from much lower precision and
recall. This imposes great challenges to training accurate NER models.

\begin{table}[tb!]
	\centering
	\caption{Existing Gazetteer Matching Performance on Open-Domain \citep{sang2003introduction, strauss2016results} and Biomedical Domain NER Datasets \citep{shang2018learning}. }
	\vspace{+0.1in}
	\begin{tabular}{|c | c | c | c | c|} 
		\hline
		\multirow{2}{*}{Metric} & \multicolumn{2}{c|}{Open-Domain} & \multicolumn{2}{c|}{Biomedical Domains} \\ 
		\cline{2-5}
		 & CoNLL03  & Tweet & BC5CDR & NCBI-Disease \\
		\hline
		Entity Types 	& 4 & 10 & 2 & 1 \\
		\hline
		F-1 			& 59.61 & 35.83 & 71.98 & 69.32\\
		Precision 		& 71.91 & 40.34 & 93.93 & 90.59 \\
		Recall 		& 50.90 & 32.22 & 58.35 & 56.15\\
		\hline
	\end{tabular}
	\vspace{-0.1in}
\label{tab:comp_distantlabel}
\end{table}

\subsection{Pre-trained Language Model} 

Pre-trained language models, such as BERT and its variants (\eg, RoBERTa \citep{liu2019roberta}, ALBERT
\citep{Lan2020ALBERT} and T5 \citep{raffel2019exploring}), have achieved
state-of-the-art performance in many natural language understanding tasks \citep{jiang2019smart}. 
These models are essentially massive neural networks based on bi-directional transformer architectures, and are trained using open-domain data in a completely unsupervised manner. The stacked self-attention modules of the transformer architectures can capture deep contextual information, and their non-recurrent structures enable the training to scale to large amounts of open-domain data. For example, the popular BERT-base model contains 110 million parameters, and is trained using the BooksCorpus~\citep{zhu2015aligning} (800 million words) and English Wikipedia (2500 million words). More importantly, many pre-trained language models have been publicly available online. One does not need to train them from scratch. When applying pre-trained language models to downstream tasks, one only needs to slightly modify the model and adapt the model through efficient and scalable stochastic gradient-type algorithms.


\section{Two-Stage Framework: \ours}

\begin{figure*}[ht]
	\centering
	\includegraphics[width=1\textwidth]{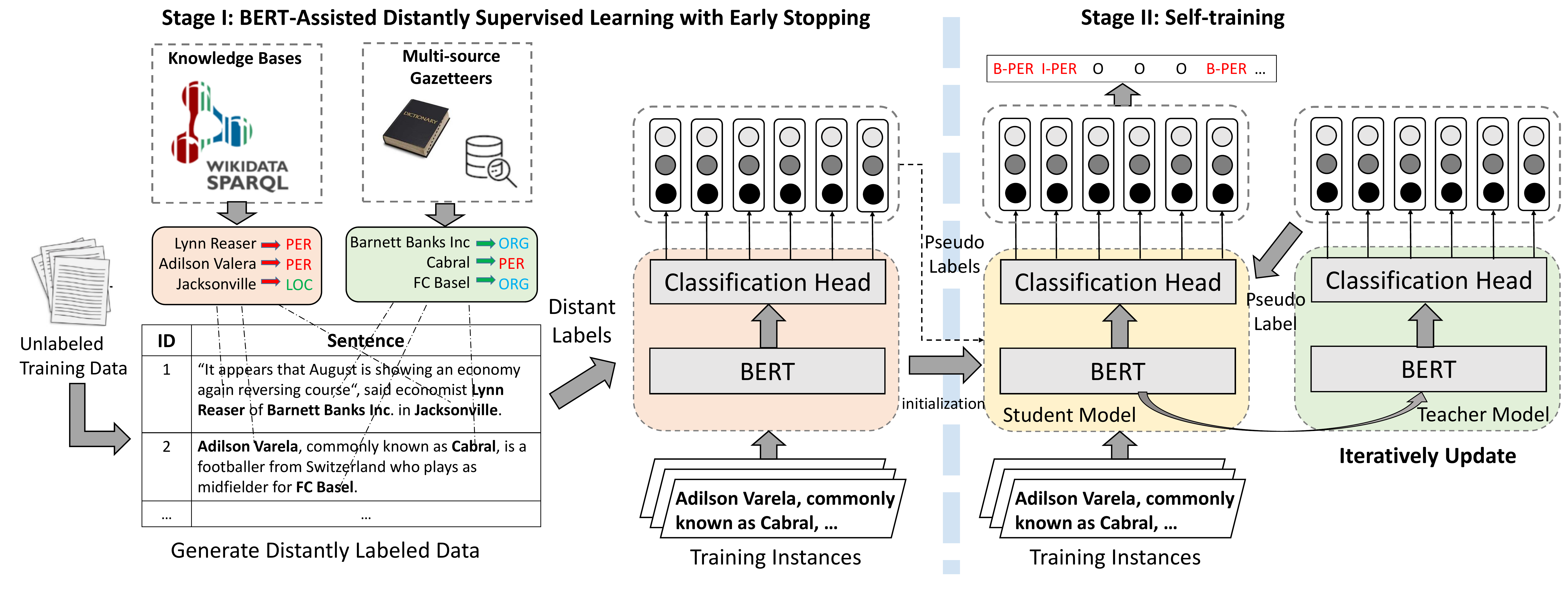}
	\vspace{-0.3in}
	\caption{The two-stage \ours framework. In Stage I, the pre-trained BERT is adapted to the distantly supervised NER task with early stopping. In Stage II, a student model and a teacher model are first initialized from the model learned in Stage I. Then the student model is trained using pseudo-labels generated by the teacher model. Meanwhile, the teacher model is iteratively updated by the early-stopped student. }
	\label{fig:FlowChart}
\end{figure*}

We introduce our proposed two-stage framework--\ours. In the first stage of \ours, we adapt the BERT model to the distantly supervised NER task. In the second stage, we use a self-training approach to improve the model fitting to the training data. We summarize the \ours framework in Figure~\ref{fig:FlowChart}.

\subsection{Stage I: BERT-Assisted Distantly Supervised Learning with Early Stopping}

\begin{figure}[ht]
	\centering
	\includegraphics[width=0.6\textwidth]{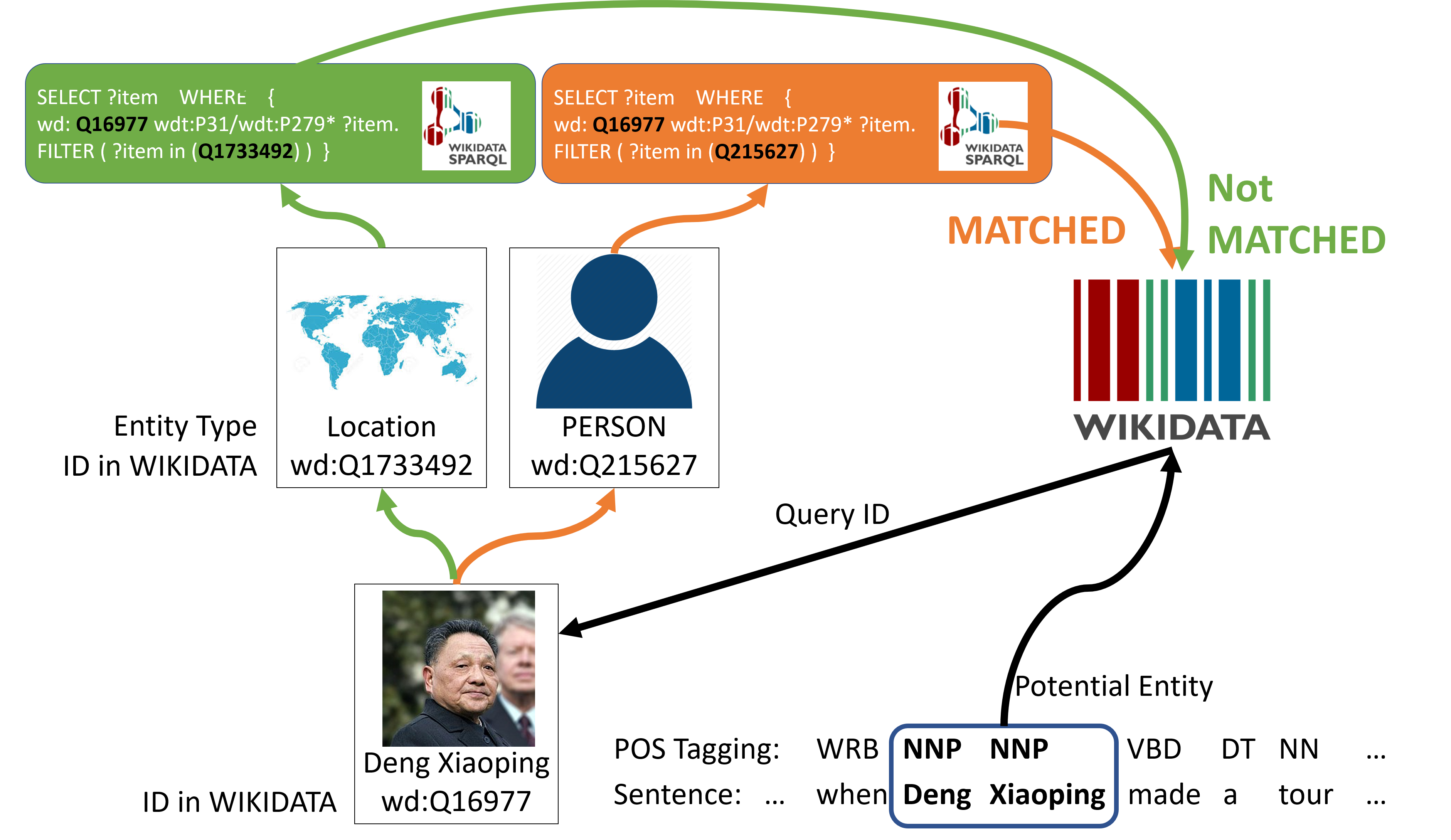}
	\vspace{-0.15in}
	\caption{Illustration of matching entities from Wikidata}
	\label{fig:wikimatch_small}
\end{figure}

Before proceeding with our proposed method, we briefly introduce how we generate distant labels for open-domain NER tasks. Our label generation scheme contains two steps: We first identify potential entities by POS tagging and hand-crafted rules. We then query from Wikidata to identify the types of these entities using SPARQL \citep{vrandevcic2014wikidata} as illustrated in Figure~\ref{fig:wikimatch_small}. We next collect gazetteers from multiple online resources to match more entities in the data \citep{sang2003introduction}. Please refer to the appendix for more technical details.

We then proceed with our proposed method. We use $f(\cdot; \theta)$ to denote the NER model parameterized by $\theta$, $f_{n,c}(\cdot; \cdot)$ to denote the probability of the $n$-th token belonging to the $c$-th class, and $\{(\bX_m,\bD_m)\}_{m=1}^M$ to denote the distantly labeled data, where $\bD_m = [d_{m,1}, ..., d_{m,N}]$ and $\bX_m = [x_{m,1}, ..., x_{m,N}]$. The NER model $f(\cdot; \theta)$ is learned by minimizing the loss over $\{(\bX_m,\bD_m)\}_{m=1}^M$: 
\begin{align}
\hat\theta = \argmin_{\theta}\frac{1}{M}\sum_{m=1}^{M} \ell(\bD_m, f(\bX_{m}; \theta)),
\label{eq:stage1}
\end{align}
where $\ell(\bD_m, f(\bX_{m}; \theta)) = \frac{1}{N} \sum_{n=1}^{N} -\log{f_{n,d_{m, n}}(\bX_{m}; \theta)}$. 

\begin{figure}[ht]
	\centering
	\includegraphics[width=0.7\textwidth]{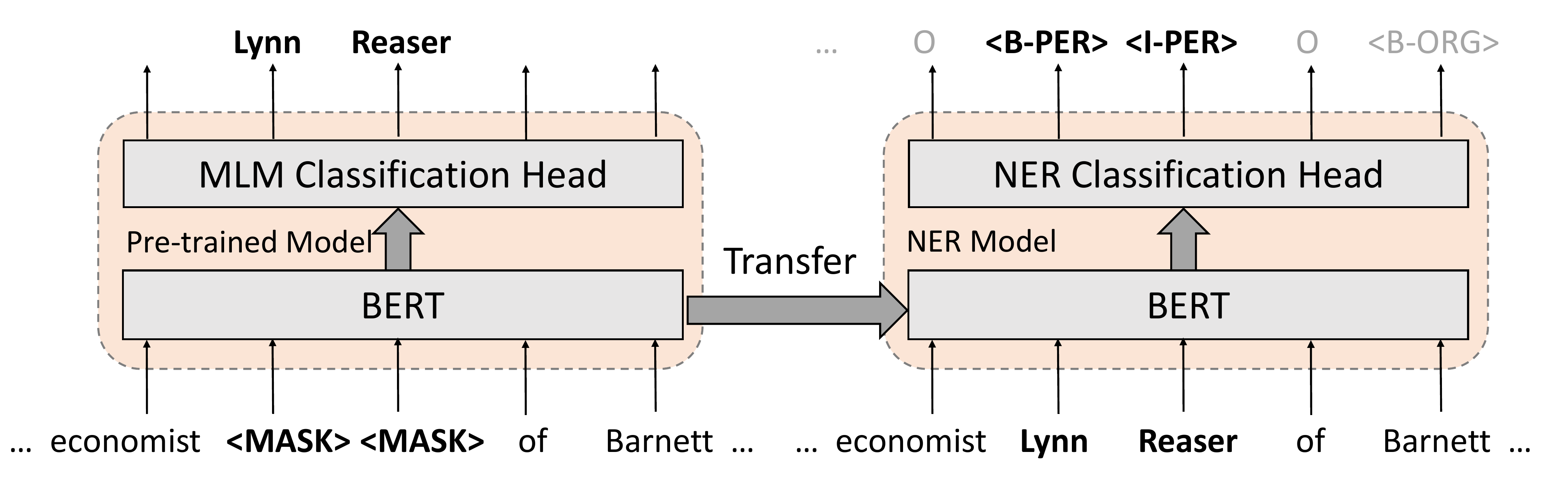}
	\vspace{-0.15in}
	\caption{Pre-trained Mask Language Model vs. NER Model}
	\label{fig:Architecture}
\end{figure}

The architecture of the NER model $f(\cdot, \cdot)$ is a token-wise NER classifier on top of a pre-trained BERT, as shown in Figure~\ref{fig:Architecture}.  The NER classifier takes in the token-wise output embeddings from the 
pre-trained BERT layers, and gives the prediction on the type for each token. 
The pre-trained BERT contains rich semantic and syntax knowledge, and yields high quality 
output embeddings. Using such embeddings as the initialization, we can efficiently adapt 
the pre-trained BERT to the target NER task using stochastic gradient-type algorithms, 
e.g., ADAM \citep{kingma2014adam,Liu2019}. Following \cite{raffel2019exploring}, 
our adaptation process updates the entire model including both the NER classification 
layer and the pre-trained BERT layers.

\begin{algorithm}[ht]
	\KwIn{$M$ unlabeled sentences, $\{\bX_m\}_{m=1}^M$; External KBs including Wikidata and multi-source gazetteers; The NER model with pre-trained BERT layers $f(\cdot; \theta^{(0)})$; The early stopping time $T_1$; The updating formula of ADAM $\mathcal{T}$.}
	\textbf{// Distant Label Generation (DLG)} 
	\begin{align*}
	\{\bD_m\}_{m=1}^M = \textrm{Matching}(\{\bX_m,\bD_m\}_{m=1}^M; \textrm{External KBs})
	\end{align*}
	\textbf{// Model Adaptation} \\
	\For{$t = 1, 2, ..., T_1$}{
		Sample a minibatch $\cB_t$ from $\{(\bX_m,\bD_m)\}_{m=1}^M$ .\\		
		Update the model using ADAM:\\
		\quad\quad\quad\quad\quad\quad$
		\theta^{(t)} = \cT(\theta^{(t-1)}, \cB_t) .
		$
	}
	\KwOut{The early stopped model: $\hat\theta = \theta^{(T_1)}$}
	\caption{Stage I:  BERT-Assisted Distantly Supervised Learning with Early Stopping}
	\label{alg:main1}
\end{algorithm} 

\begin{figure}[ht]
	\centering
	\includegraphics[width=0.6\textwidth]{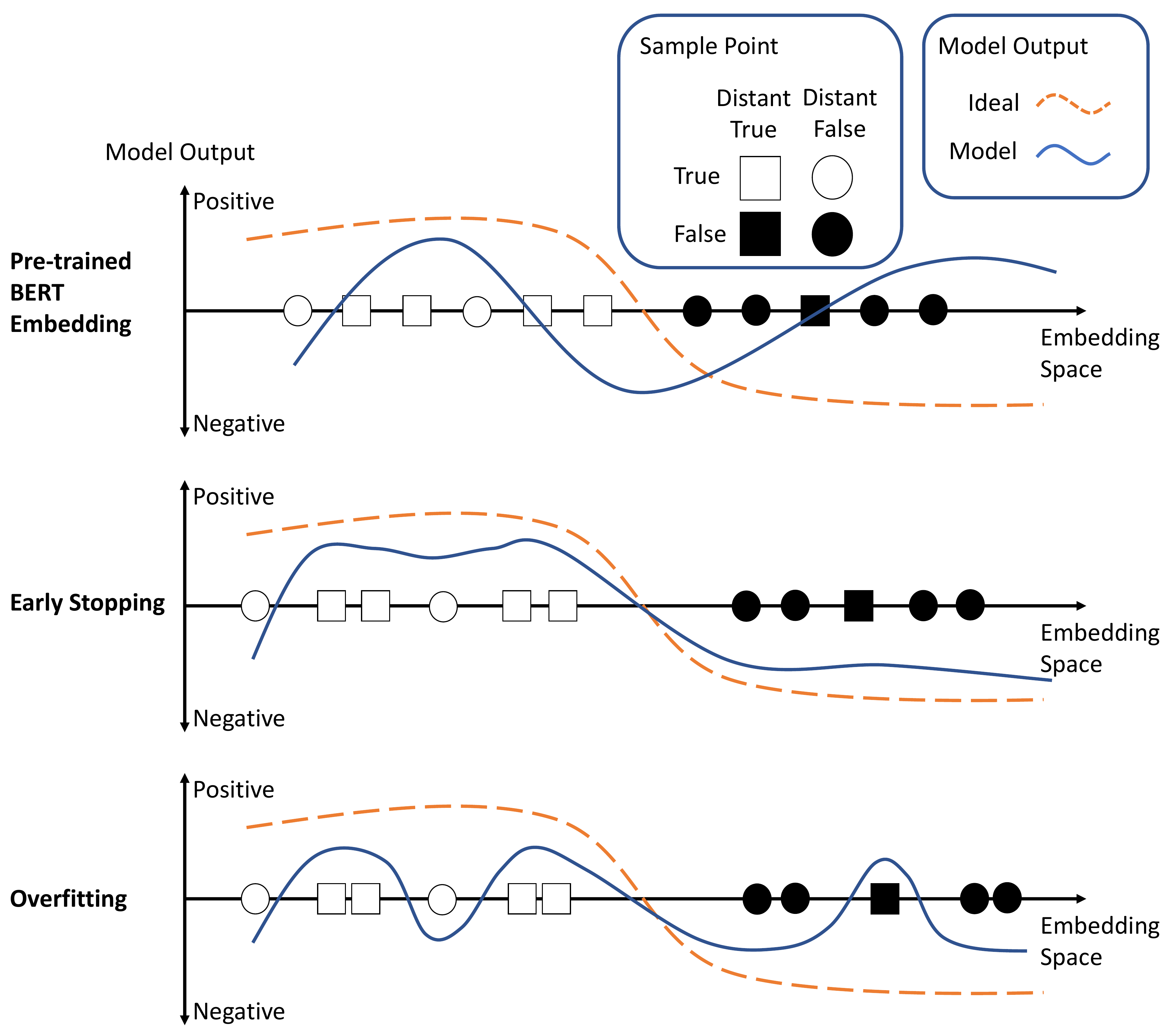}
	\vspace{-0.15in}
	\caption{
		Illustration of Stage I. 
		Top) The pre-trained semantic knowledge is transferred to the NER task;
		Middle) Early stopping leverages the pre-trained knowledge and yields better prediction;
		Bottom) Without early stopping, the model overfits the noise. 
		The token embeddings are evolving, as we update the pre-trained BERT layers.
	}
	\label{fig:overfit}
\end{figure}

Figure~\ref{fig:overfit} illustrates how the pre-trained BERT embeddings help the model adapt to distantly supervised NER tasks. We highlight that BERT is pre-trained through a masked language model (MLM) task, and is capable of predicting the missing words using the contextual information. Such a MLM task shares a lot of similarity with the NER task. Both of them are token-wise classification problems and heavily rely on the contextual information (see Figure~\ref{fig:Architecture}). This naturally enables the semantic knowledge of the pre-trained BERT to be transferred to the NER task. Therefore, the resulting model can better predict the entity types than those trained from scratch using only the distantly labeled data.

\vspace{1ex}\noindent
\textbf{Early Stopping.} 
One important strategy we use in the adaptation process is early stopping. 
Due to the large model capacity as well as the limited and noisy supervision (distant labels), our NER model can overfit the noise in distant labels and forget the knowledge of the pre-trained BERT if without any intervention. Early stopping essentially serves as a strong regularization to prevent such overfitting and improves generalization ability to unseen data.

\begin{remark}
Stage I addresses both of the two major challenges in distantly supervised NER tasks: noisy annotation and incomplete annotation. As the semantic knowledge in the pre-trained BERT is transferred to the NER model, the noise is suppressed such that the prediction precision is improved. Moreover, early stopping prevents the model from overfitting the incomplete annotated labels and further improves the recall.
\end{remark}

\subsection{Stage II: Self-Training}
We first describe a teacher-student framework of self-training to improve the model fitting, and then we propose to use high-confidence soft labels to further improve the self-training.

\subsubsection{The Teacher-student Framework}

We use $f(\cdot; \theta_{\textrm{tea}})$ and $f(\cdot; \theta_{\textrm{stu}})$ to denote teacher and student models, respectively. Given the model learned in Stage I, $f(\cdot; \hat\theta)$, one option is to initialize the teacher model and the student model as: $$\theta_{\textrm{tea}}^{(0)} = \theta_{\textrm{stu}}^{(0)} = \hat\theta,$$ and another option is 
\begin{align}\label{re-init}
\theta_{\textrm{tea}}^{(0)} = \hat\theta\quad\textrm{and}\quad\theta_{\textrm{stu}}^{(0)}=\theta_{\textrm{BERT}},
\end{align}
where $\theta_{\textrm{BERT}}$ denotes the initial model with the pre-trained BERT layers used in Stage I. For simplicity, we refer the second option to ``re-initialization''.

At the $t$-th iteration, the teacher model generates pseudo labels $\{\tilde{\bY}^{(t)}_m = [\tilde{y}_{m,1}^{(t)}, ..., \tilde{y}_{m,N}^{(t)}]\}_{m=1}^{M}$ by
\begin{align}
\tilde{y}_{m,n}^{(t)} = \argmax_{c}{f_{n,c}(\bX_m; \theta_{\textrm{tea}}^{(t)})}. 
\label{eq:pseudo}
\end{align}
Then the student model fits these pseudo-labels. 
Specifically, given the teacher model $f(\cdot;\theta_{\textrm{tea}}^{(t)})$, the student model is learned by solving 
\begin{align}
\hat\theta_{\textrm{stu}}^{(t)} = \argmin_{\theta}\frac{1}{M}\sum_{m=1}^M \ell(\tilde{\bY}_m^{(t)}, f(\bX_{m}; \theta)).
\label{eq:self_train1}
\end{align}

We then use ADAM to optimize Eq. \eqref{eq:self_train1} with early stopping. At the end of $t$-th iteration, we update the teacher model and the student model by: 
\begin{gather}
\theta_{\textrm{tea}}^{(t+1)} = \theta_{\textrm{stu}}^{(t+1)} = \hat\theta_{\textrm{stu}}^{(t)}. \notag
\end{gather}
The algorithm is summarized in Algorithm~\ref{alg:main2}.

\begin{algorithm}[ht]
	\KwIn{$M$ training sentences, $\{\bX_m\}_{m=1}^M$; The early stopped model obtained in Stage I, $f(\cdot; \hat\theta)$; The number of self-training iterations $T_2$; The early stopping time $T_3$; The updating formula of ADAM $\mathcal{T}$.}
	Initialize the teacher model and the student model:
	\[\theta_{\textrm{tea}}^{(0)} = \theta_{\textrm{stu}}^{(0)} = \hat\theta.\]
	\For{$t = 1, 2, ...T_2$}{
		$\theta_{\textrm{stu}}^{(t,0)} = \theta_{\textrm{stu}}^{(t)}.$\\
		\For{$k = 1, 2, ..., T_3$}{
			
			Sample a minibatch $\cB_k$ from $\{\bX_m\}_{m=1}^M$ .\\
			Generate pseudo-labels $\{\tilde{\bY}_m\}_{m \in \cB_k}$ by Eq. \eqref{eq:pseudo}.\\
			Update the student model:\\
			\quad\quad\quad$
			\theta_{\textrm{stu}}^{(t,k)} = \mathcal{T}(\theta_{\textrm{stu}}^{(t,k-1)},\{(\bX_m, \tilde{\bY}_m)\}_{m \in \cB_k}).
			$
		}
		
		Update the teacher and student:\\
		\quad\quad\quad\quad\quad\quad$
		\theta_{\textrm{tea}}^{(t)} = \theta_{\textrm{stu}}^{(t)} = \theta_{\textrm{stu}}^{(t,T_3)}.
		$
	}
	\KwOut{The final student model: $\theta^{(T_2)}$}
	\caption{Stage II: Self-Training}
	\label{alg:main2}
\end{algorithm}

\begin{figure}[ht]
	\centering
	\includegraphics[width=0.5\textwidth]{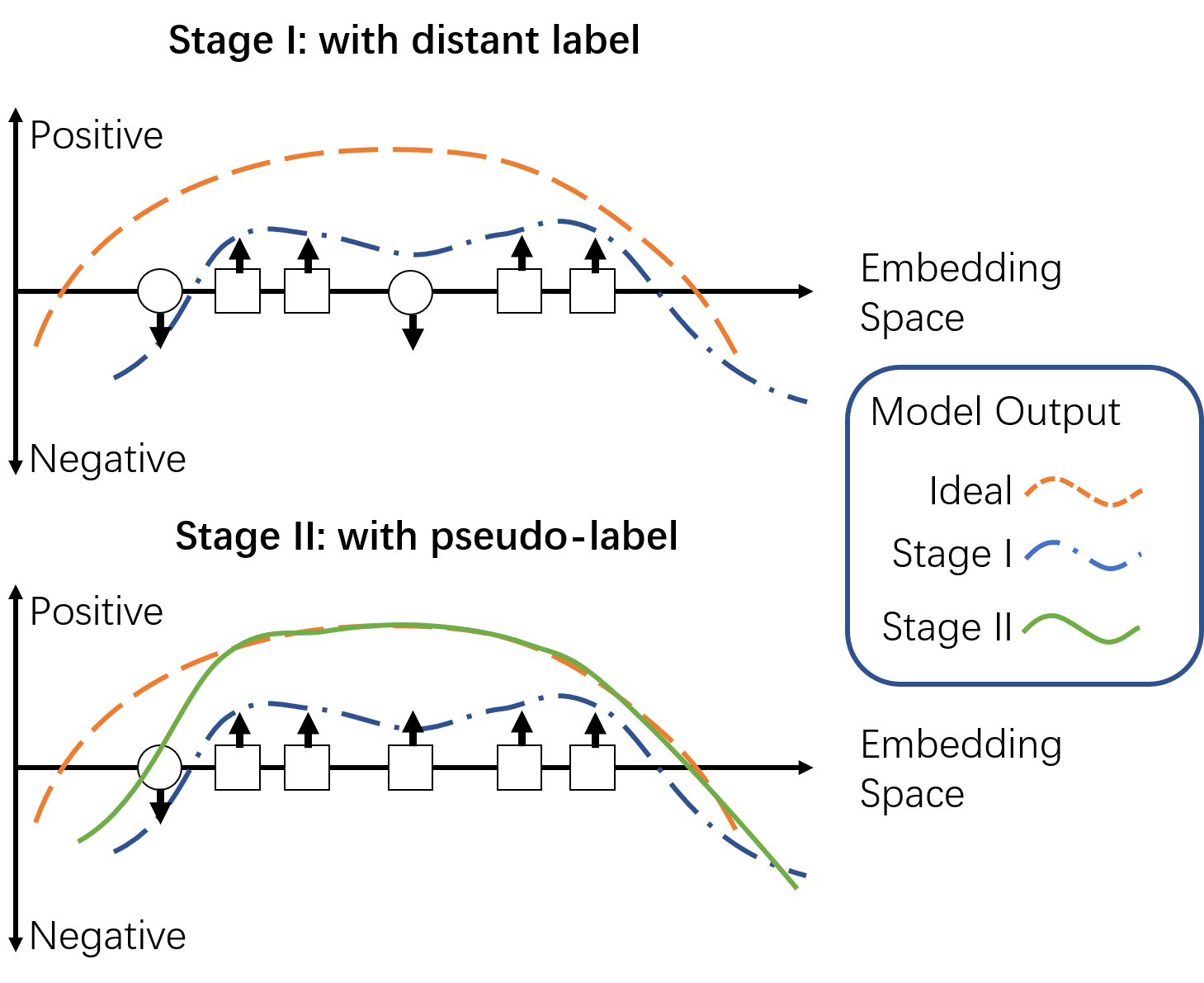}
	\vspace{-0.15in}
	\caption{Illustration of self-training. The self-training can gradually reduce the noise of the pseudo-labels and improve model fitting.
	}
	\label{fig:stage2}
\end{figure}

\begin{remark} Note that we discard all pseudo-labels from the $(t\textrm{-}1)$-th iteration, and only train the student model using pseudo-labels generated by the teacher model at the $t$-th iteration. Combined with early stopping, such a self-training approach can improve the model fitting and reduce the noise of the pseudo-labels as illustrated in Figure~\ref{fig:stage2}. 
With progressive refinement of the pseudo-labels, the student model can gradually exploit knowledge in the pseudo-labels and avoid overfitting. 
\end{remark}

\begin{remark} Our teacher-student framework is quite general, and can be naturally combined with other training techniques, e.g., mean teacher \citep{tarvainen2017mean} and virtual adversarial training \citep{miyato2018virtual}. Please refer to Section~\ref{sec:discussion} for more detailed discussions.
\end{remark}

\subsubsection{Re-weighted High-Confidence Soft Labels}
The hard pseudo-labels generated by Eq. \eqref{eq:pseudo} only keeps the most confident class for each token. To avoid losing too much information of other classes, we propose to use soft labels with confidence re-weighting. 

Recall that for the $n$-th token in the $m$-th sentence, the output probability simplex over $C$ classes  is denoted as $$[f_{n,1}(\bX_m;\theta),...,f_{n,C}(\bX_m;\theta)].$$ At the $t$-th iteration, the teacher model generates soft pseudo-labels $\{\bS_m^{(t)} = [\bs_{m,n}^{(t)}]_{n=1}^N \}_{m=1}^M$ following ~\cite{xie2016unsupervised}:
\begin{align}
\bs_{m,n}^{(t)} = [s_{m,n,c}^{(t)}]_{c=1}^{C} = \Bigg[  \frac{f_{n,c}^2(\bX_m;\theta_{\textrm{tea}}^{(t)})/p_{c}}{\sum_{c'=1}^C f_{n,c'}^2(\bX_m;\theta_{\textrm{tea}}^{(t)})/p_{c'}}\Bigg]_{c=1}^{C}
\label{eq:soft}
\end{align} 
where $p_{c} = \sum_{m=1}^M \sum_{n=1}^N f_{n,c}(\bX_m;\theta_{\textrm{tea}}^{(t)})$ calculates the unnormalized frequency of the tokens belonging to the $c$-th class. As can be seen, such a squared re-weighting step in Eq. \eqref{eq:soft} essentially favors the classes with higher confidence. The student model $f(\cdot; \theta_{\textrm{stu}}^{(t)})$ is then optimized by minimizing
\begin{align*}
\theta_{\textrm{stu}}^{(t)} &= \argmin_{\theta} \frac{1}{M} \sum_{m=1}^{M} \ell_{\rm KL}(\bS_m^{(t)}, f(\bX_{m}; \theta)),
\end{align*}
where $\ell_{\rm KL}(\cdot,\cdot)$ denotes the KL-divergence-based loss:
\begin{align}
\ell_{\rm KL}(\bS_m^{(t)}, f(\bX_{m}; \theta))=\frac{1}{N}\sum_{n=1}^N\sum_{c=1}^C - s_{m,n,c}^{(t)} \log f_{n,c}(\bX_{m}; \theta).
\label{eq:klloss}
\end{align}

\noindent \textbf{High-Confidence Selection.} 
To further address the uncertainty in the data, we propose to select tokens based on the prediction confidence. Specifically, at the $t$-th iteration, we select a set of high confidence tokens from the $m$-th sentence by
\begin{align}\label{select-token}
H^{(t)}_m = \{n : \max_{c} s_{m,n,c}^{(t)} > \epsilon \},
\end{align}
where $\epsilon\in(0,1)$ is a tuning threshold. Accordingly, the student model $f(\cdot; \theta_{\textrm{stu}}^{(t)})$ can be optimized by minimizing the loss only over the selected tokens:
\begin{align*}
\theta_{\textrm{stu}}^{(t)} 
= \argmin_{\theta} \frac{1}{M|H^{(t)}_m|}\sum_{m=1}^{M} \sum_{n\in H^{(t)}_m}\sum_{c=1}^C - s_{m,n,c}^{(t)} \log f_{n,c}(\bX_{m}; \theta).
\end{align*}
The high confidence selection essentially enforces the student model to better fit tokens with high confidence, and therefore is able to improve the model robustness against low-confidence tokens.


\section{Experiments} \label{sec:exp}
We conduct a series of experiments to demonstrate the superiority of our proposed method.
\subsection{Experimental Setup}
\subsubsection{Datasets}
We consider the following NER benchmark datasets:
(i) \textbf{CoNLL03} \citep{tjongkimsang2003conll} is a well-known open-domain
NER dataset from the CoNLL 2003 Shared Task.  It consists of 1393 English news
articles and is annotated with four entity types: person, location,
organization, and miscellaneous.
(ii) \textbf{Twitter} \citep{godin2015multimedia} is from the WNUT 2016 NER shared task. This is an open-domain NER dataset that consists of 2400 tweets (comprising 34k tokens) with 10 entity types. 
(iii) \textbf{OntoNotes5.0} \citep{weischedel2013ontonotes} contains text documents
from multiple domains, including broadcast conversation, P2.5 data and Web
data. It consists of around 1.6 millions words and is annotated with 18
entity types.
(iv) \textbf{Wikigold} \citep{balasuriya2009named} is a set of Wikipedia articles (40k tokens) randomly selected from a 2008 English dump and manually annotated with the four CoNLL03 entity types.
(v) \textbf{Webpage} \citep{ratinov2009design} is an NER dataset that contains
personal, academic, and computer science conference webpages.  It consists of
20 webpages that cover 783 entities belonging to the four types the same as CoNLL03.

For distant labels generation, we match entity types in external KBs including Wikidata corpus and gazetteers collected from multiple online sources. The data sources and matching details are described in the appendix. 

\subsubsection{Baselines}

We compare our model with different groups of baseline methods.

\noindent  $\bullet$ \textbf{KB Matching.} The first baseline performs string matching
with external KBs using the mechanism described in the
appendix. 

\noindent  $\bullet$  \textbf{Fully-supervised Methods.} We also include fully-supervised
NER methods for comparison, including: (i)
\textbf{RoBERTa-base}~\citep{liu2019roberta}---it adopts RoBERTa model with
linear layers to perform token-level prediction; (ii) \textbf{BiLSTM-CRF}~\citep{ma2016end} adopts
bi-directional LSTM with character-level CNN to produce token embeddings, which
are fed into a CRF layer to predict token labels.

\noindent  $\bullet$ \textbf{Distantly-supervised Methods.} The third group of baselines
are recent deep learning models for distantly-supervised NER,
including: (i)  \textbf{BiLSTM-CRF}~\citep{ma2016end} is trained using the distant labels matched from KBs; (ii)
\textbf{AutoNER}~\citep{shang2018learning} trains the model by assigning
ambiguous tokens with all possible labels and then maximizing the overall
likelihood using a fuzzy LSTM-CRF model; (iii)
\textbf{LRNT}~\citep{cao2019low} is the state-of-the-art model for
low-resource named tagging, which applies partial-CRFs on high-quality data with non-entity sampling. When
comparing with these distantly supervised methods, we use the same distant
labels as the training data for fair comparison.

\noindent $\bullet$  \textbf{Baselines with Different Settings}. The following methods also conduct open-domain NER under distant supervision. We remark that they use different KBs and extra training data. Therefore, we only compare with the results reported in their papers.
(i) \kalm~\citep{liu2019knowledge} augments a traditional language model with a KB and use entity type information to enhance the model.
(ii) \connet~\citep{lan2019learning} leverages multiple crowd annotation and
dynamically aggregates them by attention mechanism. It learn from imperfect annotations from multiple sources.\footnote{For \kalm and \connet model, the KB and crowd annotation are not public available, and thus we are unable to reproduce the results.}

\noindent  $\bullet$  For \textbf{Ablation Study}, we consider the following methods/tricks. 
(i) \textbf{MT}~\citep{tarvainen2017mean} uses Mean Teacher method to average model weights and forms a target-generating teacher model.
(ii) \textbf{VAT}~\citep{miyato2018virtual} adopts virtual adversarial training to smooth the output distribution to make the model robust to noise.
(iii) \textbf{Hard Label} generates pseudo-labels using Eq. \eqref{eq:pseudo}.
(iv) \textbf{Soft Label} generates pseudo-labels using Eq. \eqref{eq:soft}.
(v) \textbf{Reinitialization} initializes the student and teacher models using Eq. \eqref{re-init}.
(vi) \textbf{High-Confidence Selection} selects tokens using Eq. \eqref{select-token}.

\subsection{Experimental Results}
Our NER model use RoBERTa-base as the backbone. A linear classification layer is build up on the RoBERTa-base model. Please refer to the appendix for implementation details.
\subsubsection{Main Results}

Table~\ref{tab:main_result} presents the $F_1$ scores, precision and recall for all methods. Note that our implementations of the fully supervised NER methods attain very close to the state-of-the-art performance \citep{devlin2018bert,limsopatham2016bidirectional}. Our results are summarized as follows:

\noindent $\bullet$ For all five datasets, our method consistently achieves the best performance under the distant supervision scenarios, in $F_1$ score, precision and recall. In particular, our method outperforms the strongest distantly supervised NER baselines by $\{11.74, 21.91, 0.66,$ $14.35, 12.53\}$ in terms of $F_1$ score. These results demonstrate the significant superiority of our proposed method.

\noindent $\bullet$ The standard adaptation of pre-trained language models have already demonstrated remarkable performance. The models obtained by the Stage I of our methods outperform the strongest distantly supervised NER baselines by $\{5.87, 20.51, 0.42, 7.72, 4.01\}$ in terms of $F_1$ score. The Stage II of our methods further improves the performance of the Stage I by $\{5.87, 1.4, 0.24, 6.63, 8.52\}$. 

\noindent $\bullet$ On CoNLL03 dataset, compared with baselines which use different sources -- \kalm and \connet, our model also outperforms them by significant margins. More detailed technical comparisons between our method and them are provided in Section 5.
	
\begin{table*}[htb!]
	\caption{Main Results on Testing Set: $F_1$ Score (Precision/Recall) (in \%)}
	\label{tab:main_result}
	\vspace{-0.2in}
		\begin{center}
			\small
			\resizebox{\columnwidth}{!}{%
			\begin{tabular}{lccccc}
				\hline
				Method & CoNLL03 & Tweet & OntoNote5.0 & Webpage & Wikigold \\
				\hline 
				\textbf{Entity Types}
				& 4 & 10 & 18 & 4 & 4 
				\\
				\hline 
				KB Matching &$71.40 (81.13/63.75)$&$35.83 (40.34/32.22)$&$59.51 (63.86/55.71)$&$52.45 (62.59/45.14)$&$47.76 (47.90/47.63)$\\
				\hline 
				\multicolumn{6}{l}{\textbf{Fully-Supervised} (Our implementation)}\\
				RoBERTa &$90.11 (89.14/91.10)$&$52.19 (51.76/52.63)$&$86.20 (84.59/87.88)$&$72.39 (66.29/79.73)$&$86.43 (85.33/87.56)$\\
				BiLSTM-CRF &$91.21 (91.35/91.06)$&$52.18 (60.01/46.16)$&$86.17 (85.99/86.36)$&$52.34 (50.07/54.76)$&$54.90(55.40/54.30)$\\
				\hline 
				\multicolumn{6}{l}{\textbf{Baseline} (Our implementation)}\\
				BiLSTM-CRF &$59.50 (75.50/49.10)$&$21.77(46.91/14.18)$&$66.41 (68.44/64.50)$&$43.34 (58.05/34.59)$&$42.92 (47.55/39.11)$\\
				\pcrf &$67.00 (75.21/60.40)$&$26.10 (43.26/18.69)$&$67.18 (64.63/69.95)$&$51.39 (48.82/54.23)$&$47.54 (43.54/52.35)$\\
				\lrcrf &$69.74 (79.91/61.87)$&$23.84 (46.94/15.98)$&$67.69 (67.36/68.02)$&$47.74 (46.70/48.83)$&$46.21 (45.60/46.84)$\\
				\hline 
				\multicolumn{6}{l}{\textbf{Other Baseline} (Reported Results)}\\
				\kalm $^\dagger$ &$\hspace{-0.025in}76.00 (\hspace{0.085in}$-\!-\!-$\hspace{0.085in}/\hspace{0.085in}$-\!-\!-$\hspace{0.085in})$&-\!-\!-&-\!-\!-&-\!-\!-&-\!-\!-\\
				\connet$^\diamond$ &$75.57 (84.11/68.61)$&-\!-\!-&-\!-\!-&-\!-\!-&-\!-\!-\\
				\hline 
				\multicolumn{6}{l}{\textbf{Our \ours Framework}}\\
				Stage I &$75.61	(83.76/68.90)$&$46.61 (53.11/41.52)$&$68.11 (66.71/69.56)$&$59.11 (60.14/58.11)$&$51.55 (49.17/54.50)$\\
				\ours &${81.48} (82.05/80.92)$&${48.01}(53.16/43.76)$&$68.35 (67.14/69.61)$&${65.74} (67.37/64.19)$&${60.07} (53.44/68.58)$\\
				\hline
			\end{tabular}%
		    }
		\end{center}
	    \vspace{-0.1in}
		\emph{Note:} $^\dagger$: \kalm achieves better performance when using extra data. $^\diamond$:  \connet studies NER under a crowd sourcing setting, where the best human annotator achieves $F_1$ score at $89.51$.
\end{table*}

\subsubsection{Ablation Study}

To gain insights of our two-stage framework, we investigate the effectiveness of several components of our method via ablation study. The table \ref{tab:ablation} shows the results on both CoNLL03 and Wikigold datasets. Our results can be summarized as follows:

\vspace{0.05in}

\noindent $\bullet$ For Stage I, \textbf{Pre-trained Language Models} significantly improve both precision and recall for both datasets. Specifically, when training the NER model from scratch, the F1 scores of the output model of Stage I drop from $75.61$ to $36.66$ on CoNLL03, and from $51.55$ to $18.31$ on Wikigold. This verifies that the rich semantic and contextual information in pre-trained RoBERTa has been successfully transferred to our NER model in Stage I.

\vspace{0.05in}

\noindent $\bullet$  For Stage I, \textbf{Early stopping} improves both precision and recall for both datasets. We increase the training iterations from $900$ to $18000$ on CoNLL03 and from $350$ to $7000$ on Wikigold, and the F1 scores of the output model of Stage I drop from $75.61$ to $72.11$ on CoNLL03, and from $51.55$ to $49.68$ on Wikigold. This verifies that Early Stopping eases the overfitting and improves the generalization ability of our NER model.


\vspace{0.05in}

\noindent $\bullet$  For Stage II, \textbf{Soft labels} improve the $F_1$ score and recall on both datasets. Specifically, the $F_1$ scores and recall increase from $77.28/71.98$ to $80.18/78.84$ on CoNLL03, and from $56.90/59.74$ to $58.64/65.79$ on Wikigold. Moreover, the precision on Wikigold is also improved. This verifies that the soft labels preserve more information and yield better fitted models than those of the hard labels.

\vspace{0.05in}	

\noindent $\bullet$  For stage II, \textbf{High-Confidence Selection} improves the $F_1$ scores on both datasets. Specifically, compared with using soft labels, the $F_1$ scores and recall increase from $81.56/78.84$ to $80.18/72.31$ on CoNLL03, and from $58.64/59.74$ to $60.07/68.58$ on Wikigold. Besides, the precision on CoNLL03 is also improved. This verifies that the high-confidence labels help select data and yield more robust performance.

\vspace{0.05in}

\noindent $\bullet$ For Stage II, \textbf{Re-initialization} improves both precision and recall, only when the hard labels are adopted. We believe that this is because the hard labels lose too much information about data uncertainty, re-initializing the RoBERTa layers restores semantic and contextual information, and can compensate such loss. 

In contrast, when soft labels are adopted, \textbf{Re-initialization} deteriorates both precision and recall. We believe that this is because the soft label retains sufficient information (i.e., the knowledge transferred from RoBERTa and learned from the distant labels). As a result, re-initialization only leads to underfitting on the data. 

\begin{table}[htb!]
	\caption{Ablation Study: $F_1$ Score (Precision/Recall) (in \%)}
	\label{tab:ablation}
	\vspace{-0.1in}
	\begin{center}
		\begin{tabular}{l@{\hspace{0.05in}}c@{\hspace{0.05in}}c}
			\toprule
			Method & CoNLL03 & Wikigold \\
			\hline 
			\multicolumn{3}{l}{Stage I}\\
			\hline
			Stage I &$75.61 (83.76/68.90)$&$51.55 (49.17/54.50)$\\
			Stage I w/o pre-train &$36.66 (37.49/35.75)$&$18.31 (18.14/18.50)$\\
			Stage I w/o early stop &$72.11 (81.65/64.57)$&$49.68 (48.67/50.74)$\\
			Stage I w/ MT &$76.30 (82.92/70.67)$&$46.68 (49.82/43.91)$\\
			Stage I w/ VAT &$76.38	(82.58/71.04)$&$47.54 (50.02/45.30)$\\
			\hline 
			\multicolumn{3}{l}{Stage I + Stage II}\\
			\hline
			\ours$^\dagger$ &$77.28 (83.42/71.98)$&$56.90 (54.32/59.74)$\\
			\ours w/ soft &$80.18 (81.56/78.84)$&$58.64 (58.29/65.79)$\\
			\ours w/ soft+high conf &$81.48 (82.05/80.92)$&$60.07 (53.44/68.58)$\\
			\ours w/ reinit &$78.17 (85.05/72.31)$&$58.55 (55.31/62.19)$\\
			\ours w/ soft+reinit &$76.92(83.39/71.38)$&$54.09 (50.72/57.94)$\\
			\ours w/ MT &$77.16 (82.79/72.25)$&$57.93 (55.66/60.39)$\\
			\ours w/ VAT &$77.64 (85.62/70.69)$&$57.39 (55.05/59.41)$\\
			\bottomrule
		\end{tabular}
	\end{center}
	\vspace{-0.1in}
\emph{Note$^\dagger$:} We use $\ours$ to denote our two-stage framework using hard pseudo-labels in this table for clarity. 
\end{table}

Moreover, we also consider {\bf Multiple Re-initialization}, and observe similar results.

\begin{figure*}[!htb]
	\centering
	\begin{tabular}{ @{}c@{ }c@{ } }
		\includegraphics[width=0.5\textwidth]{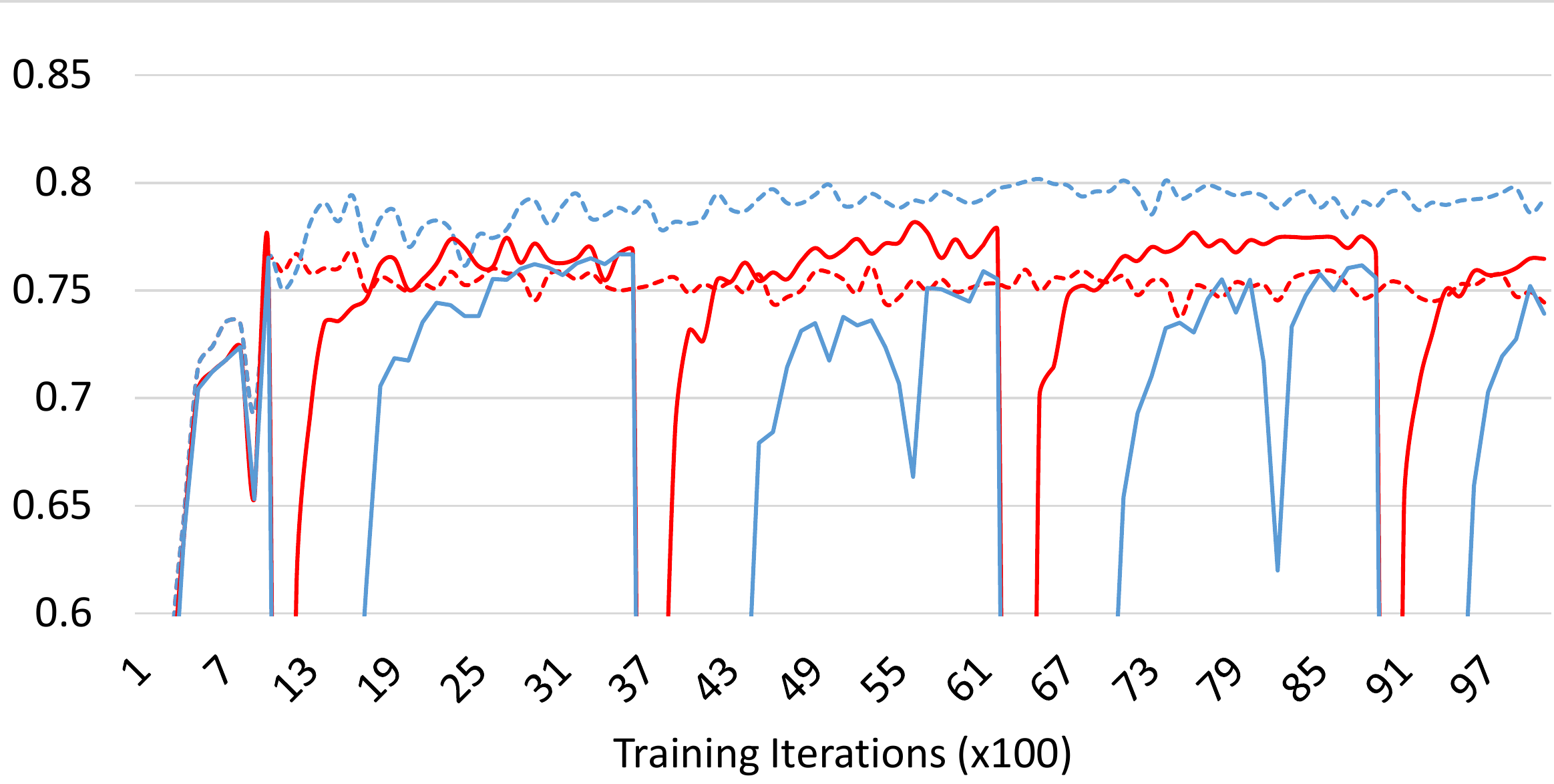} & 
		\includegraphics[width=0.5\textwidth]{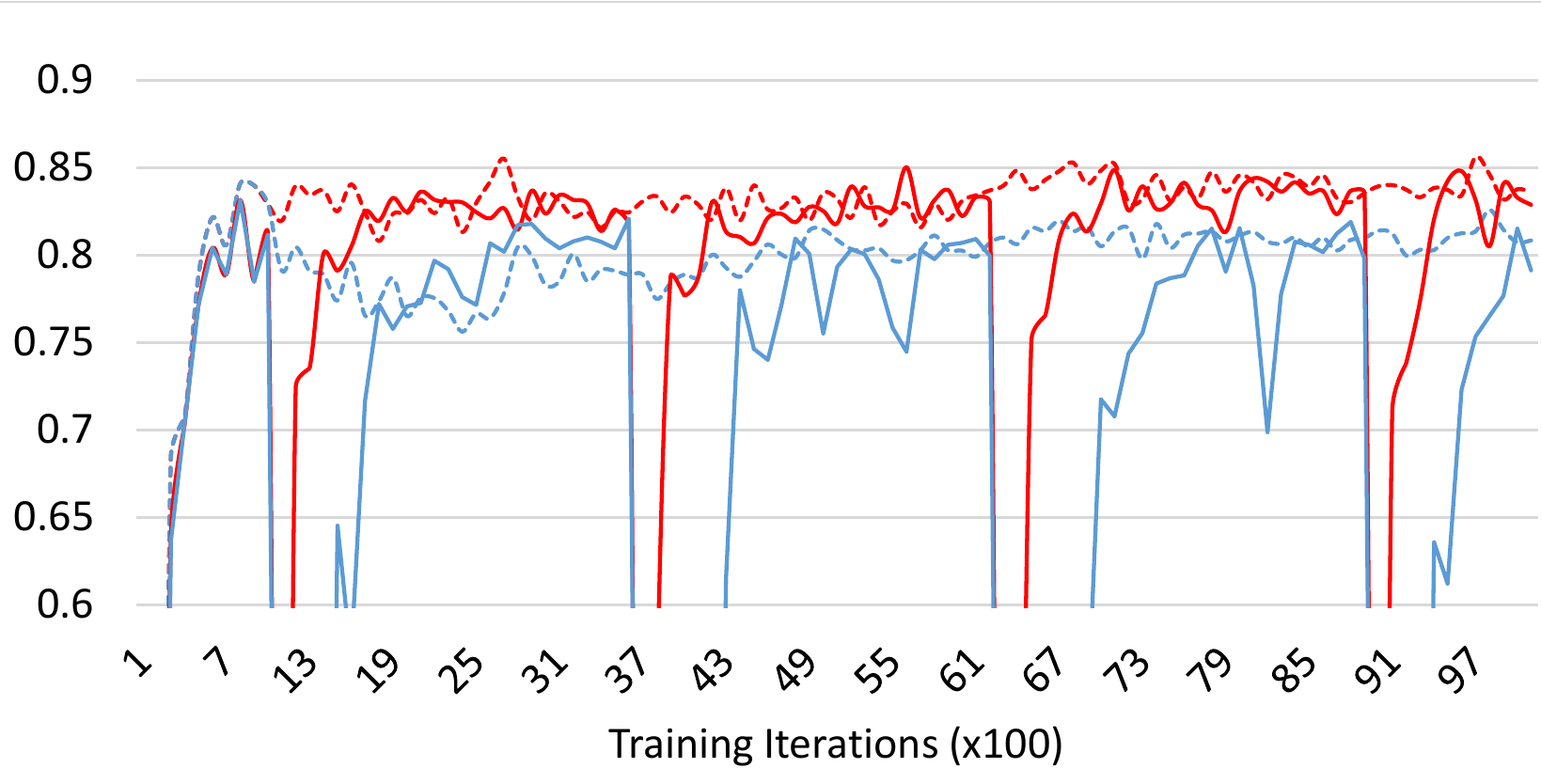} \\ 
		(a) $F_1$ score & (b) Precision 
		\vspace{+0.1in}
	\end{tabular}
	\begin{tabular}{ @{}c@{}}
		\includegraphics[width=0.5\textwidth]{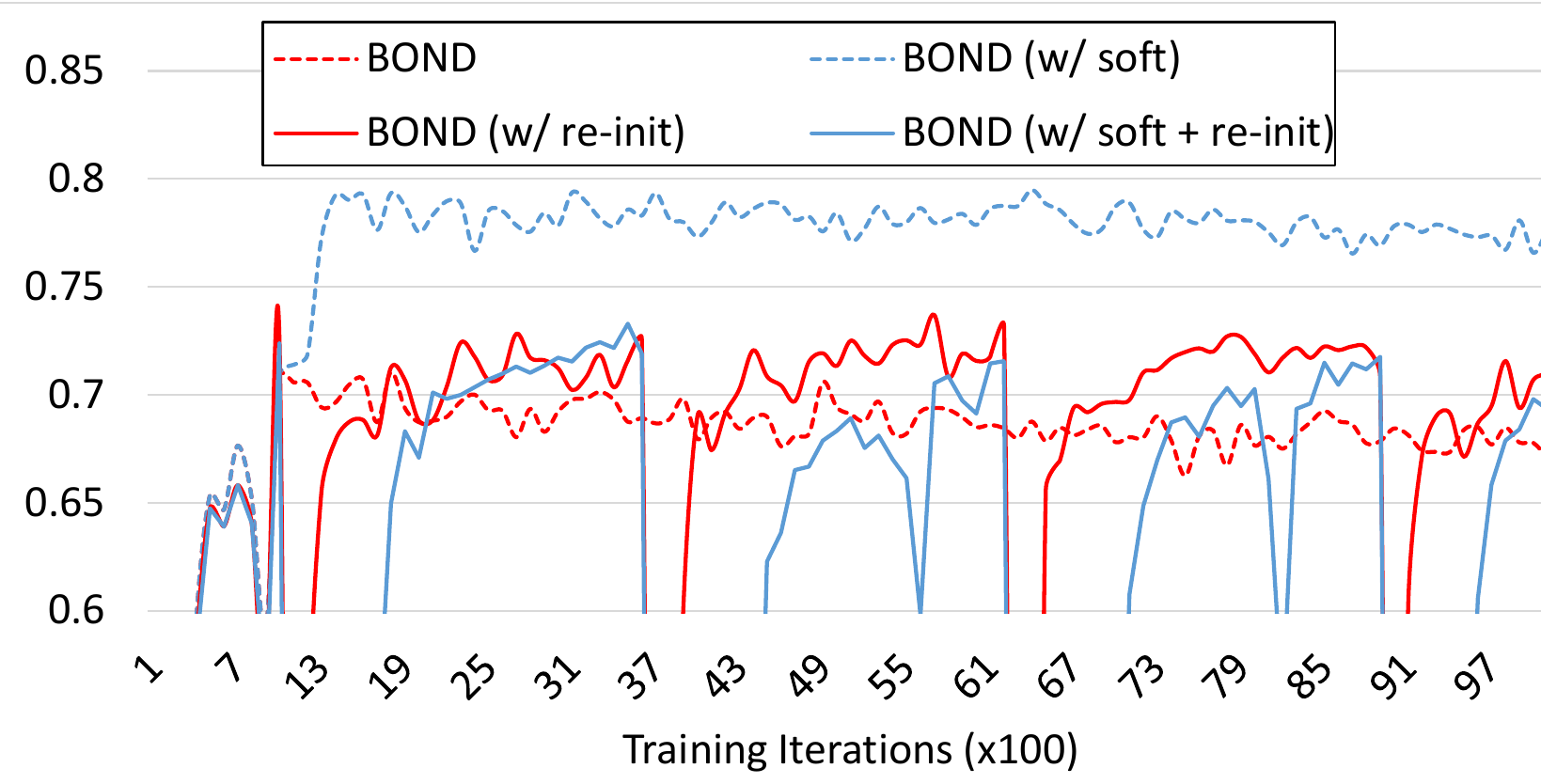} \\
		(c) Recall
	\end{tabular}
	\caption{Learning Curves of \ours, \ours (w/ reinit), \ours (w/ soft) and \ours (w/ soft + reinit)}
	\label{fig:learning_curve}
\end{figure*}

\noindent $\bullet$  \textbf{Mean Teacher} and \textbf{Virtual Adversarial Training} can be naturally integrated into our versatile teacher-student framework by adding an additional MT teacher or a VAT teacher. \textbf{VAT} marginally improves the F1 scores on both datasets. \textbf{MT} marginally improves the F1 scores on Wikigold, and deteriorates the $F_1$ scores on CoNLL03. We believe that this is because \textbf{MT} and \textbf{VAT} perform well with high quality labels, however, the labels in our NER tasks are not very precise.

\subsubsection{Parameter Study}

We investigate the effects of the early stopping time of Stage I -- $T_1$, the early stopping time of Stage II-- $T_3$, and confidence threshold $\epsilon$ for selecting tokens using CoNLL03 data. The default values are $T_1 = 900, T_3 =1800, \epsilon=0.9$. The learning curves are summarized in Figure \ref{fig:learning_curve}:

\noindent $\bullet$ Both $T_1$ and $T_3$ reflect trade-offs between precision and recall of the Stage I and Stage II, respectively. This verifies the importance of early stopping. The model performance is sensitive to $T_1$, and less sensitive to $T_3$.

\noindent $\bullet$ The recall increases along with $\epsilon$. The precision shows a different behavior: it first decreases and then increases.

\noindent $\bullet$ We also consider a scenario, where $T_3$ is allowed to tune for each iteration of the Stage II. This requires more computational resource than the setting where $T_3$ remains the same for all iterations. This can further improve the model performance to $83.49$, $84.09$, $82.89$ in terms of $F_1$ scores, precision and recall, respectively.



\begin{figure*}[!htb]
	\centering
	\begin{tabular}{ @{}c@{ }c@{ }c@{} }
		\includegraphics[width=0.32\textwidth]{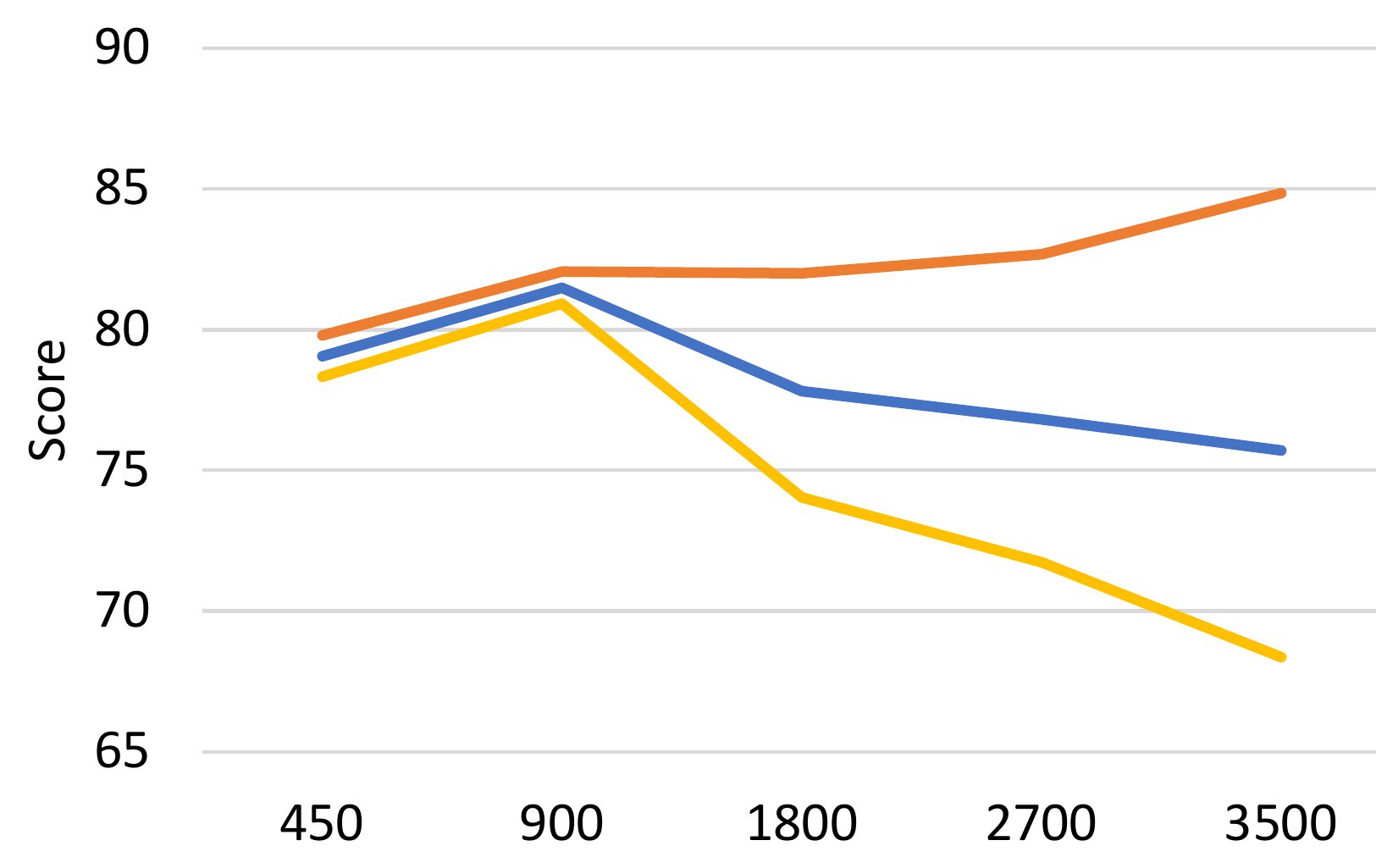} & 
		\includegraphics[width=0.32\textwidth]{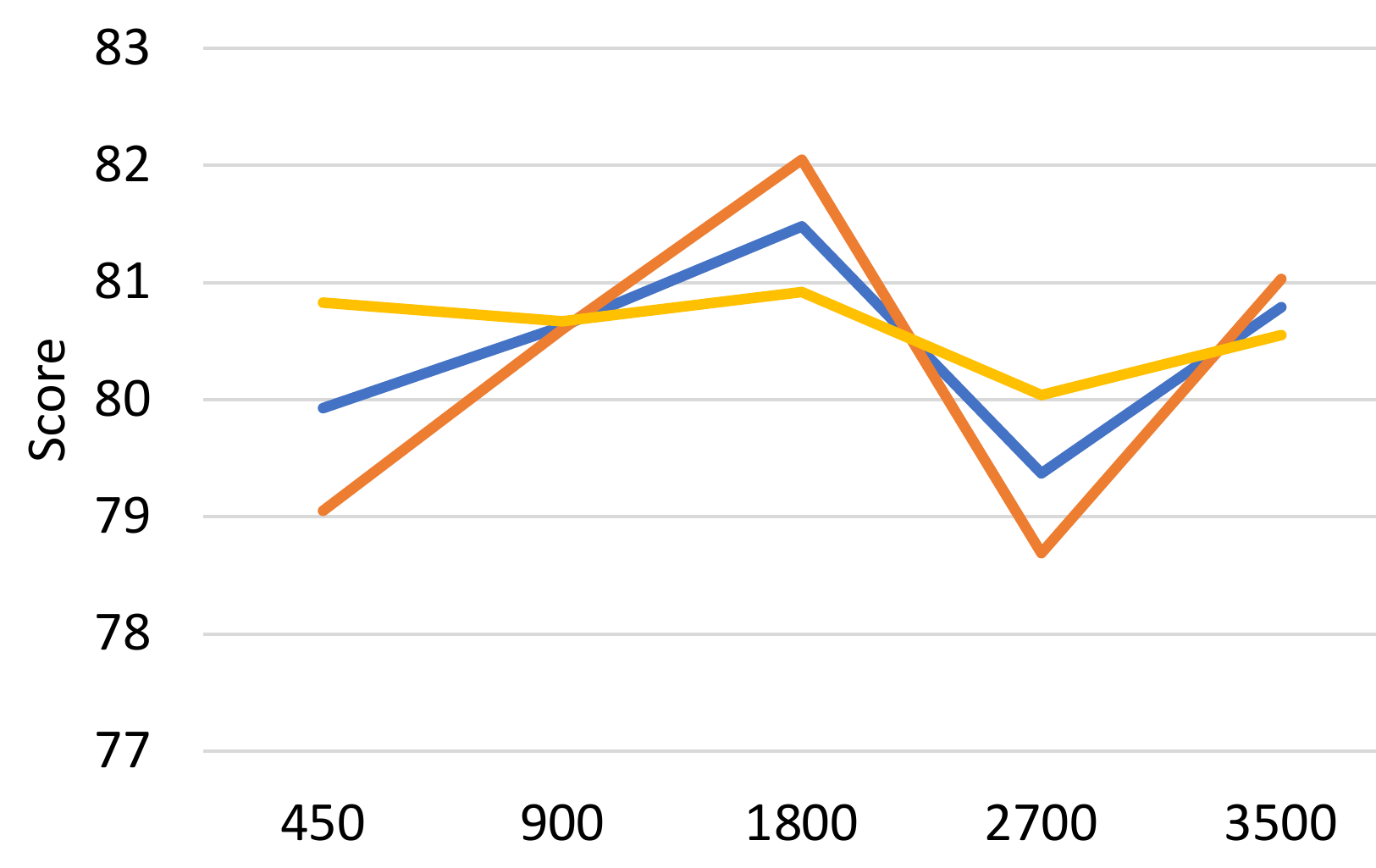} & 
		\includegraphics[width=0.32\textwidth]{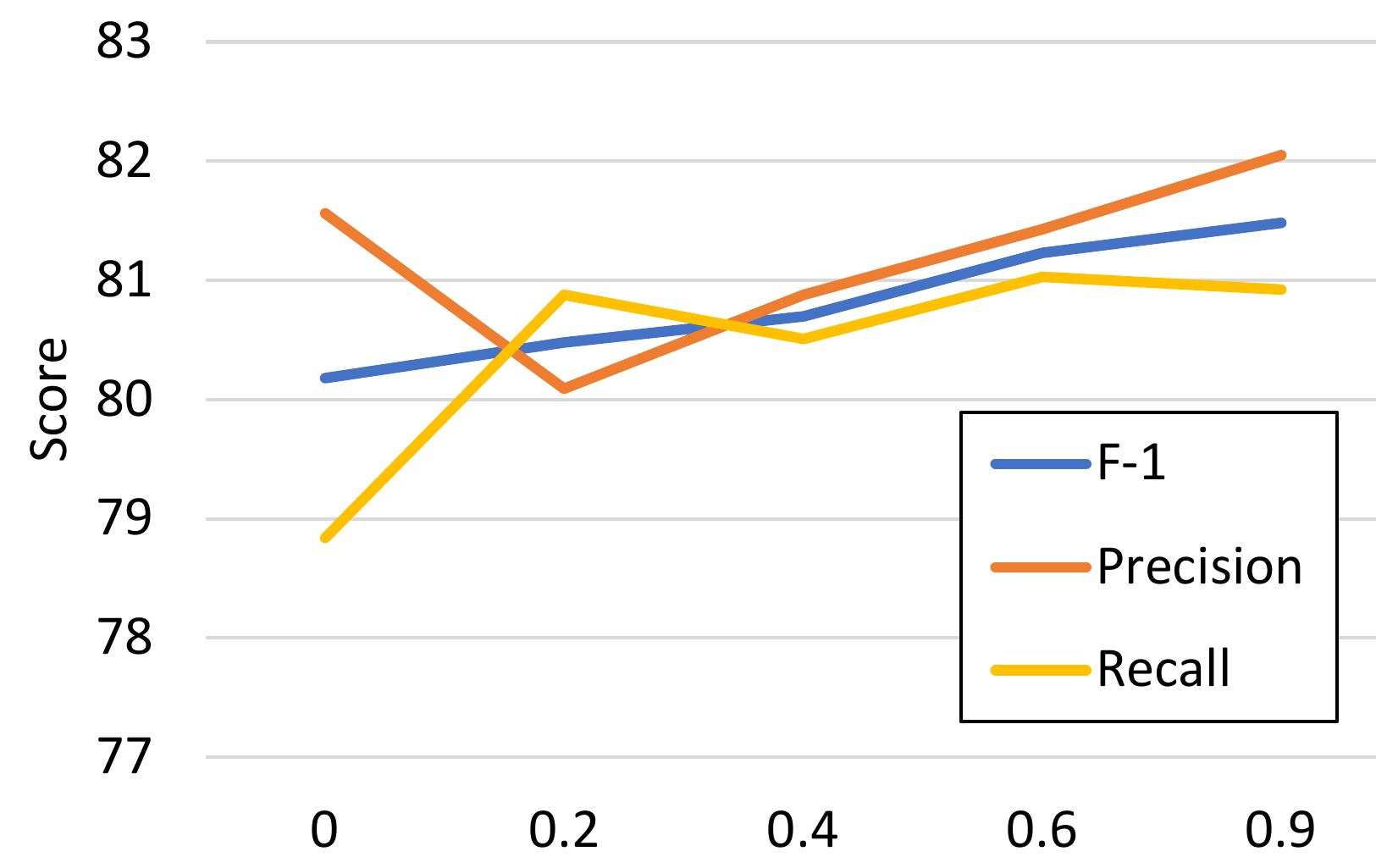} \\ 
		(a) The Early Stopping Time & (b) The Early Stopping Time & (c) The Confidence Threshold \\
		of Stage I -- $T_1$ & in Stage II -- $T_3$ &  of Stage II -- $\epsilon$
	\end{tabular}
	\caption{Parameter Study using CoNLL03: $F_1$, Precision, Recall on Testing Set (in \%)}
	\label{fig:parameter_study}
\end{figure*}

\vspace{-0.05in}
\subsubsection{Case Study and Error Analysis}

\begin{figure*}[!hbt]
	\centering
	\begin{tabular}{ @{}c@{ }c@{ }c@{} }
		\includegraphics[width=0.32\textwidth]{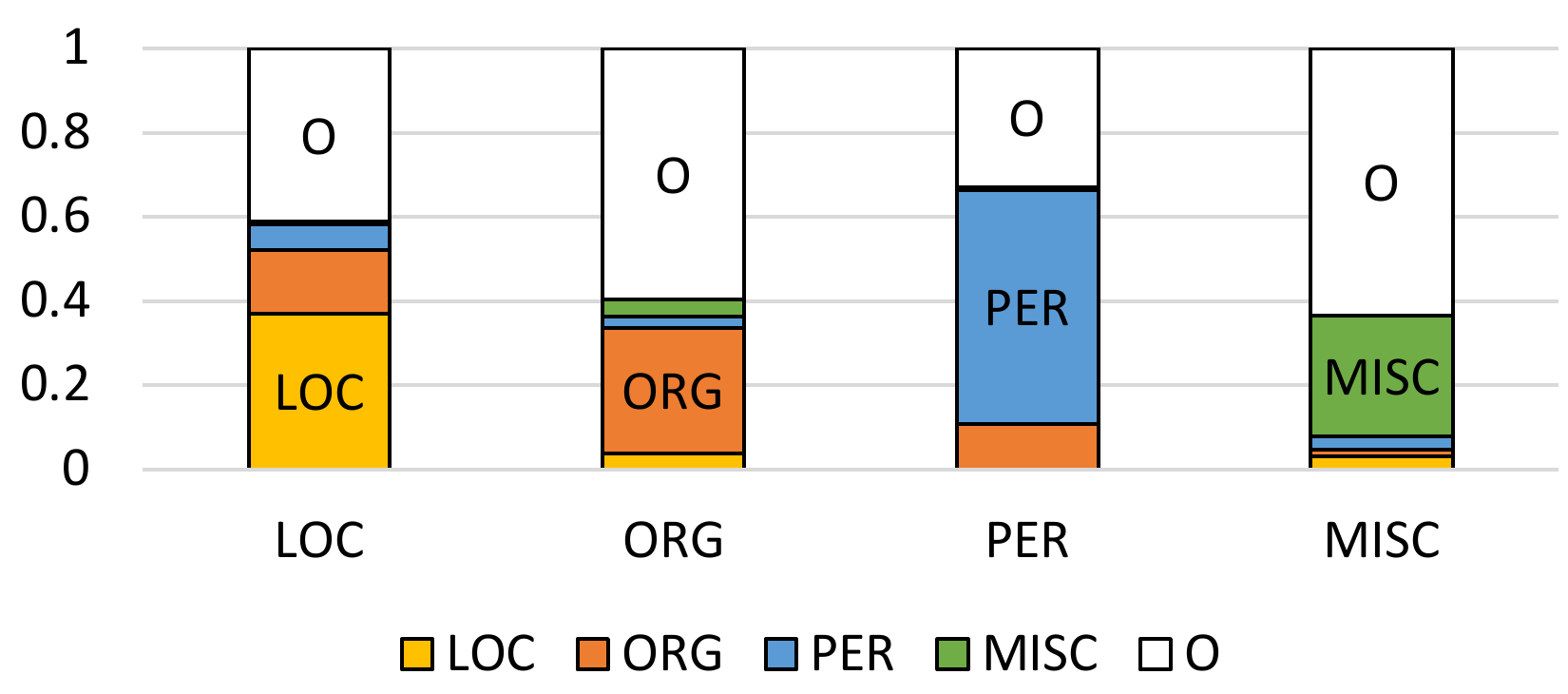} & 
		\includegraphics[width=0.32\textwidth]{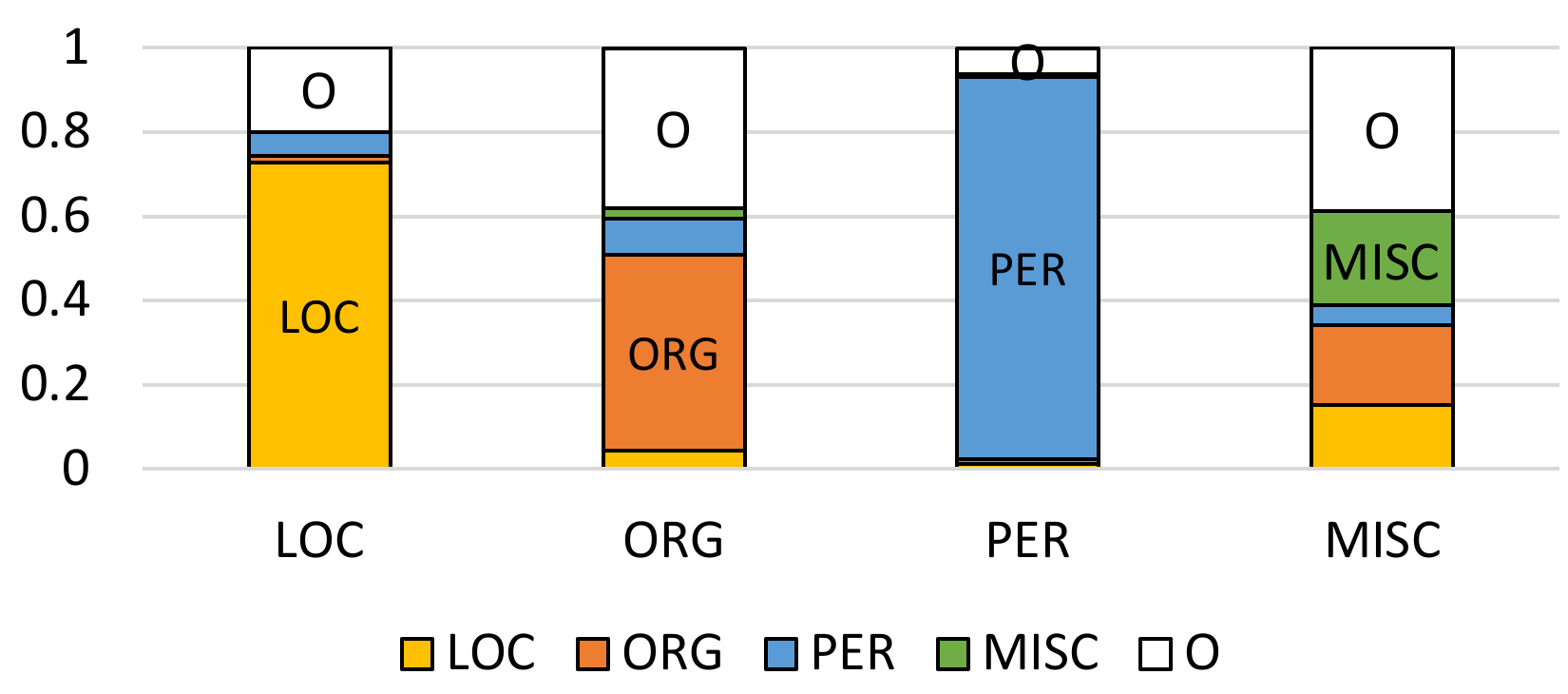} & 
		\includegraphics[width=0.32\textwidth]{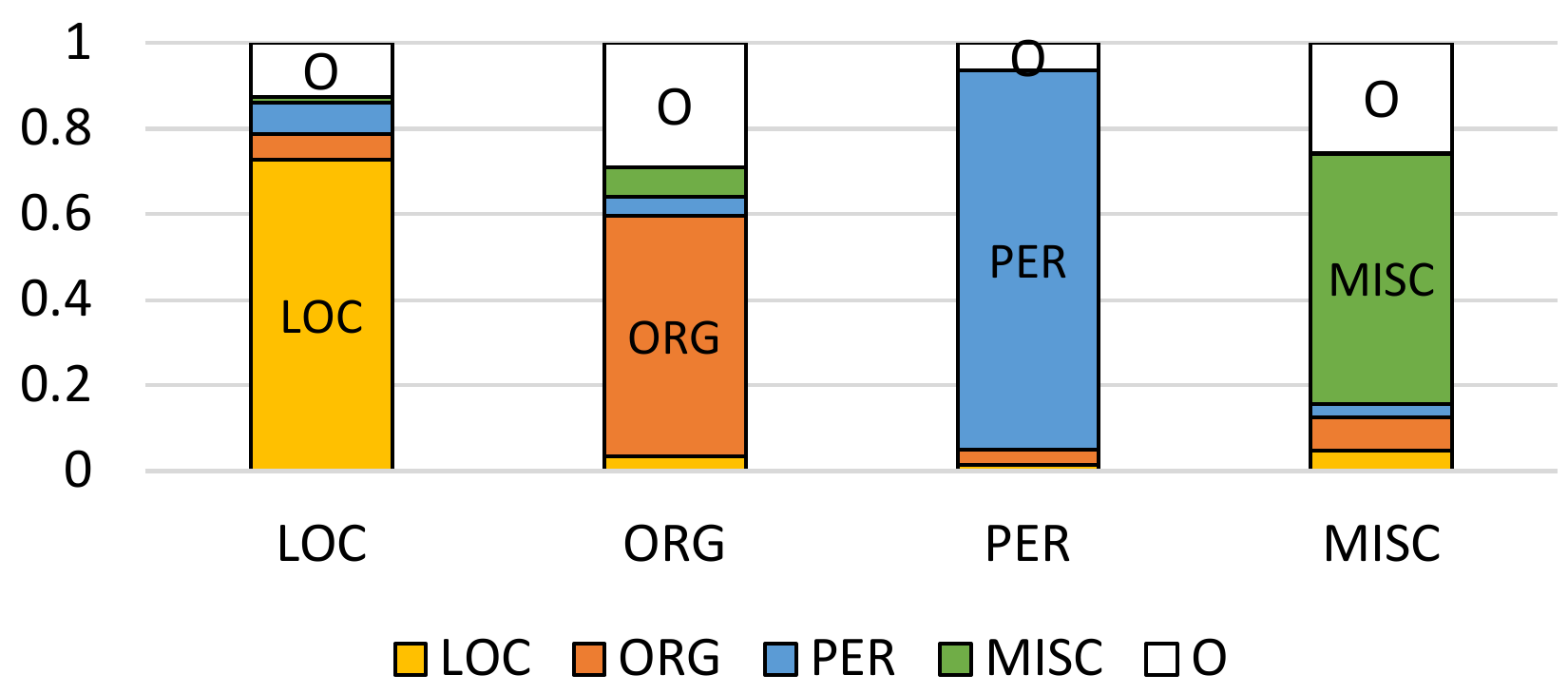} \\ 
		(a) Knowledge Base Matching & (b) Stage I & (c) Stage II
	\end{tabular}
	\caption{Recall of Knowledge Base Matching and different stages of \ours. The horizontal axis denotes the true entity type. The segments in a bar denote the portions of the entities being classified into different entity types. }
	\label{fig:hist}
\end{figure*}

To demonstrate how \ours improves the recall, we compare the prediction performance of KB matching with the output models of Stage I and Stage II using Wikigold data. Figure \ref{fig:hist} presents the bar plots of four entity types -- ``\texttt{LOC}'', ``\texttt{PER}'', ``\texttt{ORG}'' and ``\texttt{MISC}''. As can be seen, the KB matching yields a large amount of ''\texttt{O}'' (non-entity) due to its limited coverage. As a result, the recall is very low $47.63\%$. In contrast, our model of the Stage I benefits from the transferred knowledge of pre-trained RoBERTa and is able to correct some wrongly matched \texttt{O}'s to their corresponding entity types. Therefore, it enjoys a better recall $54.50\%$. Moreover, the self-training in the Stage II further improves the recall to $68.48\%$.

\section{Related Work and Discussion}
\label{sec:discussion}

Our work is related to \textbf{low-resource NER}. 
This line of research focuses on leveraging cross lingual information to 
improve the model performance. For examples, \cite{cotterell-duh-2017-low, ijcai2018-566} 
consider NER for a low resource target language. They propose to train an NER model with 
annotated language that are closely related 
to the target language. \cite{xie-etal-2018-neural} propose to use the bilingual dictionaries to tackle this challenge. More recently, \cite{DBLP:journals/corr/abs-1902-00193} propose a Bayesian graphical model approach to further improve the low resource NER performance. 

\noindent Our work is also relevant to \textbf{semi-supervised learning}, where 
the training data is only partially labeled. There have been many semi-supervised 
learning methods, including the popular Mean Teacher and Virtual Adversarial Training 
methods used in our experiments for comparison \citep{rosenberg2005semi,tarvainen2017mean,miyato2018virtual,meng2018weakly,clark2018semi}. 
Different from distant supervision, these semi-supervised learning methods usually has a partial set of labeled data. They rely on the labeled data to train an sufficiently accurate model. The unlabeled data are usually used for inducing certain regularization to further improve the generalization performance. The distant supervision, however, considers the setting with only noisy labels. Existing semi-supervised learning methods such as Mean Teacher and Virtual Adversarial Training can only marginally improve the performance, as shown in the ablation study in our experiments.

\noindent \textbf{Other related works}: \cite{liu2019knowledge} propose a language model-based method --- \kalm for NER tasks. However, their approach has two drawbacks: (i) Since they design a language model designated for NER tasks, they need to first train the language models from scratch. However, this often requires a large amount of training corpus and enormous computational resources. In contrast, \ours uses general-purpose pre-trained language models, which are publicly available online. (ii) The training of their language model is not fully unsupervised and requires token-level annotations. To address this issue, they resort to distant supervision, which yields incomplete and noisy annotations. Therefore, their language model does not necessarily achieve the desired performance.

\noindent \textbf{Larger Pre-trained Language Models}: To further improve the performance of \ours, we can use larger pre-trained language models such as RoBERTa-large~\citep{liu2019roberta} (Three times as big as RoBERT-base in our experiments) and T5~\citep{raffel2019exploring} (Thirty times larger than RoBERTa-base). These larger models contain more general semantics and syntax information, and have the potentials to achieve even better performance for NER Tasks. Unfortunately, due to the limitation of our computational resources, we are unable to use them in our experiments.

\bibliographystyle{ims}
\bibliography{refer}
\clearpage
\appendix
\section{Detailed Description of Distant Label Generation} \label{app:label_generation}

\subsection{External Knowledge Bases}
\textbf{Wikidata} is a collaborative and free knowledge base for the acquisition and maintenance of structured data. It contains over 100 million tokens extracted from the set of verified articles on Wikipedia. Wikidata knowledge imposes a high degree of structured organization. It provides a SPARQL query service for users to obtain entity relationships.

\noindent \textbf{Multi-sources Gazetteers.} For each dataset, we build a gazetteer for each entity type. Take CoNLL03 as an example, we build a gazetteer for the type \texttt{PER} by collecting data from multiple online sources including Random Name\footnote{\url{ https://github.com/dominictarr/random-name}}, US First Names Database\footnote{\url{https://data.world/len/us-first-names-database}}, Word Lists\footnote{\url{https://github.com/imsky/wordlists}}, US Census Bureau\footnote{\url{https://www2.census.gov/topics/genealogy/2010surnames/}}, German Surnames\footnote{\url{https://ziegenfuss.bplaced.net/zfuss/surnames-all.php?tree=1}}, Surnames Database\footnote{\url{https://www.surnamedb.com/Surname}} and Surname List\footnote{\url{https://surnameslist.org/}}. We build a gazetteer for the type \texttt{ORG} by collecting data from Soccer Team\footnote{\url{https://footballdatabase.com/ranking/world}}, Baseball Team\footnote{\url{https://www.ducksters.com/sports/list_of_mlb_teams.php}} and Intergovernmental Organization\footnote{\url{https://en.wikipedia.org/wiki/List_of_intergovernmental_organizations}}. We will release all gazetteers and codes for matching distant labels after the paper is accepted for publication.

\subsection{Distant Labels Generation Details}
We first find potential entities by POS tagging obtained from POS tagger, e.g., NLTK \citep{loper2002nltk}. 
We then match these potential entities by using Wikidata query service. Specifically, we use SPARQL to query the parent categories of an entity in the knowledge tree. We continue querying to the upper levels until a category corresponding to a type is found. For entities with ambiguity (e.g., those linked with multiple parent categories), we discard them during the matching process (i.e., we assign them with type \texttt{O}). The above procedure is summarized in Figure~\ref{fig:wikimatch_small}.

We then build, for each entity type in each dataset, a multi-sources gazetteer by crawling online data sources. Following the previous exact string matching methods \citep{sang2003introduction, giannakopoulos-etal-2017-unsupervised}, we match an entity with a type if the entity appears in the gazetteer for that type.  

For the unmatched tokens, we further use a set of hand-crafted rules to match entities. We notice that among the true entities, there is usually a stamp word. We match a potential entity with a type if there exists a stamp word in this entity that has frequent occurrence in that type. For example, "Inc." frequently occurs in organization names, thus the appearance of "Inc." indicates that the entity labels of words in the "XXX Inc." should be \texttt{B-ORG} or \texttt{I-ORG}).

Note that for Twitter, we do not build our own multi-sources gazetteer. We directly use the baseline system proposed in \cite{godin2015multimedia} to generate the distant labels.

\section{Baseline Settings}
For the baselines, we implement LSTM-CNN-CRF with Pytorch\footnote{\url{https://pytorch.org/}} and use the pre-trained $100$ dimension GloVe Embeddings~\citep{pennington2014glove} as the input vector. Then, we set the dimension of character-level embeddings to $30$ and feed them into a 2D convolutional neural network (CNN) with kernel width as $3$. Then, we tune the output dimension in range of $[25, 50, 75, 100, 150]$ and report the best performance. We train the model for $50$ epochs with early stopping. We use SGD with momentum with $m = 0.9$ and set the learning rate as $2\times 10 ^{-3}$. We set the dropout rate to $0.5$ for linear layers after LSTM. We tune weight decay in range of $[10^{-5},10^{-6},10^{-7},10^{-8}]$ and report the best performance.  

For other baselines, we follow the officially released implementation from the authors: (1) AutoNER: \url{https://github.com/shangjingbo1226/AutoNER};
(2) \lrcrf: \url{https://github.com/zig-k}\\\url{win-hu/Low-Resource-Name-Tagging}.

\section{Implementation Details of \ours}

All implementation are based on the Huggingface Transformer codebase \footnote{\url{https://github.com/huggingface/transformers}}. 

\subsection{Adapting RoBERTa to the NER task}

\begin{figure}[ht]
	\centering
	\includegraphics[width=0.5\textwidth]{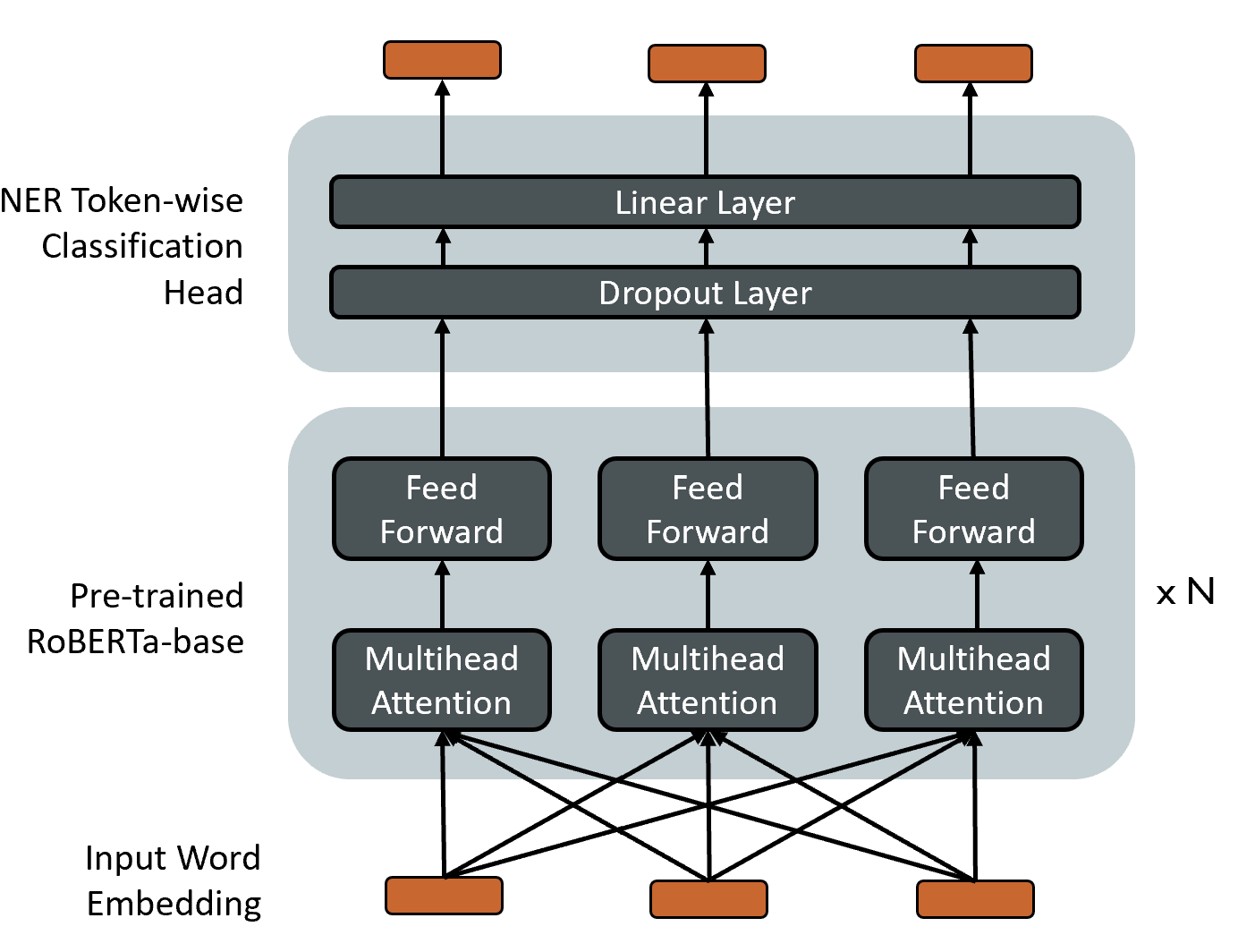}
	\vspace{-0.15in}
	\caption{The NER Model with Pre-trained RoBERTa}
	\label{fig:robert_ner_model}
\end{figure}

We choose RoBERTa-base as the backbone model of our NER model. A linear classification layer is built upon the pre-trained RoBERTa-base as illustrated in Figure~\ref{fig:robert_ner_model}. 

\subsection{Pseudo-labels Generation Details}
BERT uses WordPiece~\citep{wu2016google} for tokenization of the input text. When the teacher model predicts a set of pseudo-labels for all training data in Stage II, it assign labels for padded tokens as well. We ignore those labels in training and loss computation step by label masking.

\subsection{Parameter Settings}
There are several key parameters in our model: 1) For CoNLL03, we choose $T_1 = 900$ (about 1 epoch) and $T_3=1756$ (about 2 epochs). For Tweet, we choose $T_1 = 900$ and $T_3=900$. For OntoNotes5, we choose $T_1 = 16500$ and $T_3=1000$. For Webpage, we choose $T_1=300$ and $T_3=200$. For Wikigold, we choose $T_1=350$ and $T_3=700$. As for $T_2$, we stop training when the number of total training epochs reaches $50$ for all datasets. 2) We choose $10^{-5}$ as the learning rate for CoNLL03, Webpage and Wikigold and $2\times10^{-5}$ for OntoNotes5, Twitter, all with learning rate linear decay of $10^{-4}$. 3) We use AdamW with $\beta_1$=0.9 and $\beta_2$=0.98 as optimizer for all datasets. 4) We set $\epsilon$=0.9 for all datasets. 5) The training batch size is $16$ for all datasets except OntoNotes5.0, which uses $32$ as the training batch size. 6) For the NER token-wise classification head, we set dropout rate as $0.1$ and use a linear classification layer with hidden size $768$. For MT, we set ramp-up step as $300$ for CoNLL03, $200$ for Tweet, $200$ for OntoNotes5.0, $300$ for Webpage and $40$ for Wikigold. We choose the moving average parameters as $\alpha_1=0.99$ and $\alpha_{2}=0.995$ for all datasets. For VAT, we set the perturbation size $\epsilon_{vat}=10^{-4}$.

\subsection{Multiple Re-initialization}
Multiple Re-initialization is implemented as follows: 
In Stage II, as the performance of the student model no longer improves, we re-initialize it from the pre-trained RoBERTa-base and start a new self-training iteration. 

\subsection{Combine \ours w/ MT\&VAT}
MT\&VAT can easily combined with \ours as follows: 
During training, we update the student model by minimize the sum of weighted MT (or VAT) loss and Eq. \eqref{eq:klloss}. The weight of MT (or VAT) loss is selected in $[10,1,10^{-1},10^{-2},10^{-3}]$ using development set.

\end{document}